\documentclass[runningheads]{llncs}
\usepackage[T1]{fontenc}
\usepackage{graphicx}
\usepackage{amsmath}
\usepackage{amsfonts}
\usepackage{booktabs}
\usepackage{subcaption}
\usepackage{framed}
\usepackage[table]{xcolor}
\usepackage{multirow}
\usepackage{pifont}
\usepackage{wrapfig}
\usepackage{placeins}
\usepackage{tikz}
\usetikzlibrary{arrows.meta, positioning, calc, fit}
\definecolor{kbblue}{RGB}{40,40,160}

\newcommand{\cmark}{\ding{51}}%
\newcommand{\xmark}{\ding{55}}%

\begin{document}
\title{LELA: an LLM-based Entity Linking Approach\\ with Zero-Shot Domain Adaptation}
\titlerunning{LELA: LLM-based Entity Linking with Zero-Shot Domain Adaptation}
%
\author{Samy Haffoudhi\orcidID{0009-0008-5838-0823} \and
Fabian M. Suchanek\orcidID{0000-0001-7189-2796} \and
Nils Holzenberger\orcidID{0000-0002-0844-1391}}
\authorrunning{S. Haffoudhi et al.}
\institute{Télécom Paris, Institut Polytechnique de Paris, Palaiseau, France\\
\email{\{samy.haffoudhi, fabian.suchanek, nils.holzenberger\}@telecom-paris.fr}}
\maketitle              
\begin{abstract}
Entity linking (mapping ambiguous mentions in text to entities in a
knowledge base) is a foundational step in tasks such as knowledge graph
construction, question-answering, and information extraction. Our method, LELA,
is a modular coarse-to-fine approach that
leverages the reasoning capabilities of large language models (LLMs). It works with different target
domains, knowledge bases and LLMs, without any fine-tuning phase. 
Our experiments across various entity linking settings show that LELA is highly competitive with fine-tuned approaches, and substantially outperforms non-fine-tuned ones. 

\keywords{Entity Linking  \and Zero-shot Learning \and Large Language Models \and Information Extraction.}
\end{abstract}

\section{Introduction}

Entity linking (EL), also known as Entity Disambiguation, is the task of mapping ambiguous mentions of entities in natural language text to reference entities in a knowledge base (KB).
For instance, the input can be a sentence in which the mention of an entity has already been identified, as in this example:\\[-2mm]

\texttt{France hosted the 2024 Olympics in \underline{Paris}.\\[-2mm]}

\noindent The knowledge base contains each entity, associated with a short textual description, e.g.:\\[-2mm]

\ttfamily
\setlength{\tabcolsep}{4pt} 
\begin{tabular}{ll}
Paris (city)& Capital city of France\\
Paris (novel)& 1897 novel by Emile Zola\\
France & Country in Europe\\
\end{tabular}
\rmfamily
\ \\[2mm]
The goal is to map the mention of \underline{Paris} to the correct entity in the KB (here: the first one). Entity Linking~\cite{logeswaranZeroShotEntityLinking2019} assumes that the mention has already been identified. Thus, it is part of the broader task of End-to-End Entity Linking~\cite{decaoAutoregressiveEntityRetrieval2021,lela-demo}, which consists of the automated identification of the mention and the linking.

Entity Linking (in both variants) is an important preprocessing step in tasks such as information extraction~\cite{martinez-rodriguezInformationExtractionMeets2020}, knowledge-based question answering~\cite{weltyComparisonHardFilters2012}, and knowledge graph completion~\cite{jiKnowledgeBasePopulation,machine-knowledge-survey}. Most of the work on entity linking focuses on general knowledge
graphs such as Wikidata~\cite{vrandecicWikidataNewPlatform2012}, DBpedia~\cite{auerDBpediaNucleusWeb2007} or Yago~\cite{yago-4-5}. However, in real-world applications, the knowledge base is often proprietary or domain-specific, as in the legal or biomedical domain, or inside a company.

\begin{table}
    \caption{Classification of entity linking settings (``training'' refers to fine-tuning after initial pre-training). 
    \textbf{LELA} is in the \textbf{true zero-shot} setting.}
    \centering
    \begin{tabular}{lccc}
    \toprule
    & Domain & Knowledge Base & Documents \\
    \midrule
    Classical & test=train & test=train & test$\neq$train \\
    Zero-shot & test$\approx$train & test$\neq$train & test$\neq$train \\
    \textbf{True zero-shot} & test$\neq$train & test$\neq$train & test$\neq$train \\
    \bottomrule
    \end{tabular}
    \label{tab:el_settings}
    \vspace{-\baselineskip}
\end{table}

Therefore, several entity linking methods use training.
In \textbf{classical EL} (Table~\ref{tab:el_settings}),
the models are trained on the same domain and same KB as the one used during testing, and generalize only to new documents. In \textbf{zero-shot EL}~\cite{logeswaranZeroShotEntityLinking2019}, the set of entities seen during training is different from the one seen during testing. However, even this setting, as defined in~\cite{logeswaranZeroShotEntityLinking2019}, requires labeled entity linking training data for fine-tuning from related domains and document sources. 
Fine-tuning is not just computationally expensive, but also requires high-quality labeled data. Thus, to apply entity linking to a new domain (say, a company wants to disambiguate chemical products), one would first have to label hundreds of mentions. This task cannot be outsourced easily, as it requires navigating specialized knowledge bases and involves privacy restrictions. Hence, the quest that we tackle in this paper is \textbf{true zero-shot EL}, i.e., entity linking that generalizes also across domains.
Several approaches target such true zero-shot scenarios~\cite{dingEntGPTLinkingGenerative2024,liuOneNetFineTuningFree2024,zhouGenDeciderIntegratingNone2024,yeLLMEntityDisambiguator}. However, even these require hand-chosen Chain-of-Thought examples and/or strongly domain-specific candidate generator approaches, limiting their generalizability. Furthermore, their performance lags behind the specialized approaches. 

In this paper, we propose LELA, a true zero-shot, fine-tuning-free method for entity linking. Our key idea is that reasoning at inference time can compensate for the absence of annotated data during training time. LELA works in a coarse-to-fine fashion: It (1) generates an initial entity-candidate set, (2) selects top-$k$ candidates through pointwise re-ranking, and (3) passes them to a reasoning LLM for final candidate selection.
This approach is modular, LLM-, reranker-, and retriever-agnostic, and can hence be used with both open-source and proprietary models. As LLMs become ever more of a commodity, the performance of LELA thus only stands to gain. 
We show that  LELA systematically outperforms the true zero-shot methods, and even beats state-of-the-art classical and zero-shot approaches on multiple datasets.
Our contributions are the following:
\begin{itemize}
    \item A formalization of the \textbf{true zero-shot} entity linking setting (Table~\ref{tab:el_settings}), distinguishing it from prior zero-shot formulations.
    \item \textbf{LELA}, a modular coarse-to-fine pipeline that achieves true zero-shot EL by replacing the standard supervised paradigm with inference-time reasoning.
    \item Extensive experiments across five EL benchmarks spanning diverse domains and settings, with controlled baselines and ablations showing that LELA's gains arise from the integration of reasoning-based selection in the pipeline, rather than from individual components or backbone choice.
\end{itemize}

\section{Related Work}

\label{sec:related-work}

\subsection{Entity Linking}
Before the blossoming of machine learning and deep learning, many entity linking approaches (such as AIDA~\cite{hoffartRobustDisambiguationNamed2011}, SOFIE~\cite{suchanek_sofie}, or DBpedia Spotlight~\cite{mendes2011dbpedia}) were training-free, but were bound to a given KB (often YAGO~\cite{suchanekYagoCoreSemantic2007} or DBpedia~\cite{auerDBpediaNucleusWeb2007}).

Today, methods that work on any input KB are usually embedding-based: 
BLINK~\cite{wuScalableZeroshotEntity2020},
AcroBERT~\cite{chenGLADISGeneralLarge2023},
ExtEnD~\cite{barbaExtEnDExtractiveEntity2022}, ReFinED~\cite{ayoolaReFinEDEfficientZeroshotcapable2022}, VERBALIZED~\cite{ruckerEvaluatingDesignDecisions2025}, the approaches in~\cite{fevryEmpiricalEvaluationPretraining2020,yamadaGlobalEntityDisambiguation2022}, as well as the different zero-shot entity linking methods evaluated on the ZESHEL benchmark~\cite{logeswaranZeroShotEntityLinking2019}: ReS~\cite{xuReadandSelectFrameworkZeroshot2023}, Uni-MPR and Bi-MPR~\cite{tangBidirectionalMultiparagraphReading2021}, and E-repeat~\cite{yaoZeroshotEntityLinking2020}.
GENRE~\cite{decaoAutoregressiveEntityRetrieval2021} is a generative approach 
and mGENRE~\cite{decaoMultilingualAutoregressiveEntity2022} is its multilingual extension. 
Although these methods can technically operate over different knowledge bases, they are not true zero-shot approaches, because they require supervised training on the target KB or closely aligned domains to achieve good performance. LELA obviates this requirement, and we show in our experiments that it still performs competitively with these approaches.

Several LLM-based approaches have also been proposed. 
The approaches in~\cite{vollmersContextualAugmentationEntity2025,xinLLMAELLargeLanguage2024} use LLMs for contextual augmentation. ARTER \cite{liLeveragingPowerLarge2025} categorizes mentions and routes them to either a low-computational entity linker or a reasoning-LLM.
Others \cite{wangDeepMELMultiAgentCollaboration2025,shiGenerativeMultimodalEntity2024,qiUniMELUnifiedFramework2024} focus on the multimodal setting.
GenDecider~\cite{zhouGenDeciderIntegratingNone2024} uses an LLM fine-tuned on the reranking task on ZESHEL using LoRA~\cite{huLoRALowRankAdaptation2021}. 
OneNet~\cite{liuOneNetFineTuningFree2024} is a pipeline approach of LLM-based candidate filtering, entity linking, and hallucination tackling. 
EntGPT~\cite{dingEntGPTLinkingGenerative2024} prompts the model to augment the mention before selecting the best candidate entity and~\cite{yeLLMEntityDisambiguator} uses in-context learning~\cite{brownLanguageModelsAre2020a} for candidate selection.
Related prompting strategies have been studied for entity resolution, where
records are matched against records rather than against a knowledge base
\cite{wangMatchCompareSelect,zeakisAvengEREnsemblingFineTuning}.

All of these LLM-based approaches leverage task- or domain-specific candidate retrieval, fine-tuning, or complex prompt engineering, which limits their applicability in true zero-shot settings. %
LELA, in contrast, requires no fine-tuning or domain-specific components, and is thus truly zero-shot. Still, our experiments show that it outperforms these approaches across multiple datasets. 

\subsection{LLM-based Reranking}

LLMs have recently been explored as general-purpose rerankers in information retrieval:
\begin{itemize}
    \item pointwise approaches compute a score for each query-document pair~\cite{pipitoneZELOELOinspiredTraining2025,zhangQwen3EmbeddingAdvancing2025},
    \item pairwise approaches compare the relevance of a pair of documents~\cite{qinLargeLanguageModels2024,luoPRPGraphPairwiseRanking2024},
    \item listwise approaches process a list of documents and output them ranked by their relevance~\cite{wangJinarerankerv3LastNot2025,liuCoRankingCollaborativeRanking2025}.
\end{itemize}
Other methods are setwise~\cite{zhuangSetwiseApproachEffective2024}, tournament-inspired~\cite{chenTourRankUtilizingLarge2025}, efficiency-centered~\cite{dedovJointRankRankLarge2025}, as well as attention-pattern-based~\cite{chenAttentionLargeLanguage2025} or conditional perplexity-based~\cite{shiLongCodeZipCompressLong2025}. Some approaches make use of reasoning token generation~\cite{abdallahDeARDualStageDocument2025,liuReasonRankEmpoweringPassage2025,zhuangRankR1EnhancingReasoning2025}, and study its impact on performance~\cite{jedidiDontOverthinkPassage2025}.

This body of work primarily targets document and passage retrieval. In our work, we show that pointwise LLM-based reranking, followed by reasoning-driven selection, can be effectively adapted to the entity linking setting without task-specific fine-tuning.

\section{Methodology}

\subsubsection{Task Definition.}
We define entity linking using 4 random variables: (1)~the domain $D$ (e.g. News, Biomedical etc), (2)~the set of relevant entities $E$, (3)~the text document $T$, and (4)~the set of mentions $M$. We posit that
\begin{itemize}
    \item the set of relevant entities $E$ depends on the domain $D$,
    \item the text document $T$ depends on the domain $D$, and
    \item the set of mentions $M$ depends on the document $T$ and the set of entities $E$.
\end{itemize}

\noindent The task of entity linking is that of linking the mentions $m \in M$ in a text document $T$ to their entities in $E$. We distinguish the following settings (Table~\ref{tab:el_settings_details_here}):
In classical EL, one has access to documents sampled from a distribution conditioned on the test entities and test domain. Consequently, the relevant entities are identical at training and test time, but not the documents.
In zero-shot EL, the entities are distinct between train and test, but are sampled from the same domain. In the true zero-shot setting (which we tackle in LELA), none of the distributions are conditioned on the test domain.

\begin{table}
\caption{The different EL settings
associated with the samples from the test domain that are available for training. Mentions are always conditioned on the available document and entities, and so do not appear in the table.
$x \sim X$ means that $x$ is a realization of the random variable $X$.}
\centering
\small
\begin{tabular}{llll}
\toprule
Setting & Domain $D$ & Entity set $E$ & Text document $T$ \\
\midrule
Classical & $d_{\textrm{train}} = d_{\textrm{test}}$ & $e_{\textrm{train}} = e_{\textrm{test}}$ & $ t_{\textrm{train}} \sim T | D = d_{\textrm{test}}$ \\
Zero-shot & $d_{\textrm{train}} = d_{\textrm{test}}$ & $e_{\textrm{train}} \sim E | D = d_{\textrm{test}}$ & $ t_{\textrm{train}} \sim T | D = d_{\textrm{test}}$ \\
True zero-shot & $d_{\textrm{train}} \sim D$ & $e_{\textrm{train}} \sim E | D = d_{\textrm{train}}$ & $ t_{\textrm{train}} \sim T | D = d_{\textrm{train}}$ \\
\bottomrule
\end{tabular}
\label{tab:el_settings_details_here}
\vspace{-\baselineskip}
\end{table}


\begin{figure}[t]
\centering
\begin{tikzpicture}[
  >={Stealth[length=2mm]},
  font=\footnotesize,
  node distance=0pt,
  data/.style={rectangle, draw=black!75, align=left, inner sep=4pt, text width=6.4cm},
  fin/.style={data, very thick},
  proc/.style={rectangle, rounded corners=2pt, draw=black!55, fill=black!4,
               align=center, inner sep=3pt, text width=4.4cm},
  arr/.style={->, thick},
  ctx/.style={->, thick, rounded corners=3pt, draw=violet!70!black},
  men/.style={->, thick, rounded corners=3pt, draw=red!75!black},
  elab/.style={font=\scriptsize, inner sep=1.5pt},
  dot/.style={circle, inner sep=0pt, minimum size=2.6pt},
]
\node[draw=violet!70!black, thick, rounded corners=3pt, inner sep=4pt,
      align=center, font=\small] (sent)
  {France hosted the 2024 Olympics in \textcolor{red!75!black}{\textbf{[Paris]}}.};

\node[data, below=3.5mm of sent] (kb) {%
  \textbf{Knowledge Base}\\[1pt]
  \textcolor{kbblue}{\textbf{Paris (city)} - Capital city of France}\\
  \textcolor{kbblue}{\textbf{Paris (novel)} - 1897 novel by Zola}\\
  \textcolor{kbblue}{\textbf{Paris (Texas)} - City in Texas}\\
  \textcolor{kbblue}{\textbf{France} - Country in Europe}\\
  \textcolor{kbblue}{\dots}%
};
\node[proc, below=3.5mm of kb] (cg) {%
  \textbf{Candidate Generation}\\[0.5pt]
  {\scriptsize\textcolor{black!60}{BM25 / dense}}%
};
\node[data, below=3.5mm of cg] (ret) {%
  \textbf{Retrieved Candidates}\\[1pt]
  \textcolor{kbblue}{\textbf{Paris (city)} - Capital city of France}\\
  \textcolor{kbblue}{\textbf{Paris (novel)} - 1897 novel by Zola}\\
  \textcolor{kbblue}{\textbf{Paris (Texas)} - City in Texas}%
};
\node[proc, below=3.5mm of ret] (rr) {%
  \textbf{Candidate Reranking}\\[0.5pt]
  {\scriptsize\textcolor{black!60}{pointwise reranker}}%
};
\node[data, below=3.5mm of rr] (rank) {%
  \textbf{Reranked Candidates}\\[1pt]
  \textcolor{kbblue}{\textbf{Paris (Texas)} - City in Texas}\hfill 0.88\\
  \textcolor{kbblue}{\textbf{Paris (city)} - Capital city of France}\hfill 0.61\\
  \textcolor{kbblue}{\textbf{Paris (novel)} - 1897 novel by Zola}\hfill 0.44%
};
\node[proc, below=3.5mm of rank] (sel) {%
  \textbf{LLM-based Selection}\\[0.5pt]
  {\scriptsize\textcolor{black!60}{top-$k$ cutoff, \texttt{<think>}\dots}}%
};
\node[fin, below=3.5mm of sel] (ans) {%
  \textbf{Answer}\hspace{2ex}%
  \textcolor{kbblue}{\textbf{Paris (city)}}%
};

\draw[arr] (kb)   -- (cg);
\draw[arr] (cg)   -- (ret);
\draw[arr] (ret)  -- (rr);
\draw[arr] (rr)   -- (rank);
\draw[arr] (rank) -- (sel);
\draw[arr] (sel)  -- (ans);

\coordinate (lx) at ($(sent.west)+(-0.5,0)$);
\coordinate (rx) at ($(sent.east)+(0.5,0)$);

\draw[violet!70!black, thick, rounded corners=3pt]
  (sent.west) -- (lx) -- (sel.west -| lx);
\draw[ctx] (rr.west  -| lx) -- node[elab, above, text=violet!70!black] {full text} (rr.west);
\draw[ctx] (sel.west -| lx) -- node[elab, above, text=violet!70!black] {full text} (sel.west);
\node[dot, fill=violet!70!black] at (rr.west -| lx) {};

\draw[men] (sent.east) -- (rx) -- (cg.east -| rx)
           -- node[elab, above, text=red!75!black] {mention} (cg.east);
\end{tikzpicture}
\caption{Overview of the LELA approach, which (1) generates an initial entity-candidate set from the mention (red), (2) selects top-$k$ candidates through pointwise re-ranking, and (3) passes them to a reasoning LLM for final candidate selection. Here the reranker scores the wrong entity highest; the selection LLM recovers the correct one by reasoning over the candidates jointly.}
\label{fig:overview}
\vspace{-\baselineskip}
\end{figure}

\subsubsection{Approach.} LELA uses a three-stage \textit{coarse-to-fine} pipeline, as illustrated in Figure~\ref{fig:overview}:
Candidate entities are generated~\cite{sevgiliNeuralEntityLinking2022,wuScalableZeroshotEntity2020}, then ranked, and then selected. Beyond the text $T$, mention $m$, and KB of entities $E$, LELA requires only a parameter $k \in \mathbb{N}^+$ as input, for the top-$k$ cutoff of ranked candidates.

\subsubsection{Candidate Generation.} The goal of the candidate generation step is to identify the  entities 
${C_m \subseteq E}$ that could be relevant to $m$. For this step, LELA is retriever-agnostic, and can work with any method that is capable of pre-selecting a subset of relevant entities. In practice, any generation method that requires domain-specific training would jeopardize LELA's goal of being fine-tuning-free. Hence, we primarily consider BM25~\cite{robertsonRelevanceWeightingSearch1988} and dense retrieval, but also experiment with prior-based dictionaries. 
Note that this stage is always required in practice, as candidates are not given in a real deployment; some of the benchmarks we consider pre-supply a candidate set to enable a fair comparison across methods. We therefore use the provided candidates on ZESHEL, GLADIS and ZELDA, and generate our own on ESCO (Section~\ref{sec:esco}) and WikilinksNED Unseen Mentions (Appendix~\ref{sec:unseen_mentions}).

\newcommand{\rerankingvariable}{r}
\subsubsection{Candidate Reranking.} Each candidate $c \in C_m$ is then given a score $s(c) \in \mathbb{R}$ by pointwise reranking.
While any off-the-shelf reranker can be used, we opt for instructed rerankers, as the candidate entity reranking task is not typically covered during reranker training. Rerankers based on generative language models typically take as input a query $q$, a document $d$, and a reranking-instruction $\rerankingvariable$. They then compute a score as:
\begin{align}
\text{score}(q,d)
&= \frac{e^{\operatorname{logit}(\text{yes}\mid \rerankingvariable,q,d)}}
    {e^{\operatorname{logit}(\text{yes}\mid \rerankingvariable,q,d)} + e^{\operatorname{logit}(\text{no}\mid \rerankingvariable,q,d)}} \\
&= \operatorname{softmax} \left( \begin{smallmatrix} 
    \operatorname{logit}(\text{yes}\mid \rerankingvariable,q,d) \\ 
    \operatorname{logit}(\text{no}\mid \rerankingvariable,q,d) 
\end{smallmatrix} \right)_0
\end{align}
\noindent The score can also be expressed as a single-logit binary classifier\footnote{\url{https://huggingface.co/Qwen/Qwen3-Reranker-0.6B/discussions/3}}.
In our case, we want to obtain a score for each candidate $c \in C_m$, and hence, following the formalism of information retrieval, we set $q=T$ (marking the mention $m$), $\rerankingvariable$ as an entity disambiguation instruction (see Appendix~\ref{sec:prompts}), and $d$ as the name of $c$ followed by its description. We then identify the top-$k$ scored candidates, $C_m^{(k)}$.

\subsubsection{Candidate Selection.}
\label{sec:selection} 
To select the best candidate from $C_m^{(k)}$, we construct an LLM prompt that consists of (1)~a system instruction that asks the model to return the index of the entity that is most likely mentioned in the text (shown in Figure~\ref{fig:prompt}, Appendix~\ref{sec:prompts}), (2)~input text $T$ with the mention $m$ surrounded in square brackets, 
and (3)~the candidates as a newline-separated list of triples, each consisting of a zero-based index, an entity name, and an entity description. Optionally, the first list item is ``0. None of the candidates''. We use self-consistency~\cite{wangSelfConsistencyImprovesChain2023}, sampling different candidate selection reasoning paths, with majority voting, to determine the final candidate $c^* \in C_m^{(k)}$.

\subsubsection{Implementation.} LELA
uses vLLM\footnote{\url{https://docs.vllm.ai/}}~\cite{kwonEfficientMemoryManagement2023} for LLM-inference.
It is implemented as a Python library and can be used easily in spaCy pipelines as an \mbox{entity-linking component}\footnote{\url{https://spacy.io/}}.

\section{Experiments}

We evaluate LELA on various entity linking settings, using five different benchmark datasets.
Our evaluation spans four complementary entity linking settings: (i) domain-transfer zero-shot, (ii) domain-specific with \texttt{None}-linking, (iii) acronym disambiguation without entity descriptions across specialized domains, and (iv) classical Wikipedia-based benchmarks with strong supervised baselines. 

\subsubsection{LELA Configuration.} 
We report results for LELA using the open-source \texttt{Magistral-Small-2509}~\cite{mistral-aiMagistral2025} reasoning LLM.  We use the \texttt{Qwen3-Reranker-4B} model~\cite{zhangQwen3EmbeddingAdvancing2025} for reranking. We run our experiments on one or two \texttt{A100 40GB} GPUs based on the selected LLM's size.
For dense retrieval, we use \texttt{Qwen3-Embedding-4B}~\cite{zhangQwen3EmbeddingAdvancing2025} along with flat FAISS indexes~\cite{douzeFaissLibrary2025}. BM25 retrieval is implemented using the \texttt{bm25s} Python library\footnote{\url{https://bm25s.github.io/}}. 
Across all experiments, we use the same configuration: the selection LLM receives the top $k=10$ reranked candidates, over which we sample $k_{sc}=10$ reasoning traces for self-consistency. The number of candidates retrieved \emph{before} reranking (64 or 200, depending on the benchmark) is not a tuned parameter, as recall increases monotonically with it, and it is thus bounded only by computational cost. As we are interested in true zero-shot EL, we do not perform any hyperparameter tuning, and discuss other values of $k$, self-consistency samples and rerankers in Section~\ref{sec:analysis}.
We also report results for the mixture-of-experts (MoE)~\cite{shazeerOutrageouslyLargeNeural2017} \texttt{Qwen3-30B-A3B}, the small \texttt{Qwen3-4B}~\cite{yangQwen3TechnicalReport2025}, and the MoE \texttt{gpt-oss-20b}~\cite{openaiGptoss120bGptoss20bModel2025} LLMs.

\subsubsection{Competitors.} We compare LELA to all state-of-the-art entity linking methods that have been previously evaluated on the considered benchmarks.
The same method can be zero-shot for one dataset and true zero-shot for another one.
We also report results for \textit{LLM-SFT}, a true zero-shot baseline obtained by fine-tuning an LLM for candidate selection on Wikipedia mentions (see Appendix~\ref{sec:baseline}).
Unlike LELA, this baseline relies on transfer through parameter adaptation rather than inference-time reasoning, allowing us to compare the two paradigms directly.

\subsection{Zero-Shot Scenario}

\subsubsection{Benchmark.} ZESHEL~\cite{logeswaranZeroShotEntityLinking2019} is a benchmark
for evaluating EL methods in a zero-shot setting, built with
documents from Wikia\footnote{\url{https://www.wikia.com/}}, a Wikipedia-like Web site for
specific subjects such as fictional universes or sports. ZESHEL
consists of 8 Wikia domains for training, 4 for validation, and 4 for testing. 
The test has a total of 10,000 mentions. 
For each mention, the dataset provides 64 candidates retrieved by BM25. Entities
correspond to Wikia pages, where the entity name is the title of the page and
the entity description \mbox{is its content.}
The BM25-based candidate retrieval has a recall of 68\% on
the test set.
Following previous work \cite{logeswaranZeroShotEntityLinking2019,xuReadandSelectFrameworkZeroshot2023,liEffectiveFewShotNamed2022}, we consider the \textit{normalized accuracy}
as the evaluation metric, which is defined as the accuracy on the subset of
mentions for which the correct entity is among the candidates.
For LELA, we use the provided candidates and the first three sentences of the entity page as the entity description. 

\subsubsection{Results.} As Table~\ref{tab:zeshel_results} shows, LELA ranks first on the benchmark. It achieves the highest performance across all test domains in the true zero-shot setting, improving the previous best reported macro-averaged normalized accuracy by 8.84 percentage points. On half of the domains, it outperforms even the non-true zero shot approaches, improving the \textit{Lego} and \textit{YuGiOh} domains normalized accuracy by 1.62 and 5.06 percentage points respectively. 
Category-specific results are given in Appendix~\ref{sec:category_specific}.

\begin{table}
  \centering
  \caption{Normalized accuracy on the test set of ZESHEL. * means reported in
\cite{zhouGenDeciderIntegratingNone2024}, ** means reported in~\cite{xuReadandSelectFrameworkZeroshot2023}, *** means reported in~\cite{liEffectiveFewShotNamed2022}, - means not reported. Best results within each group are bold, best overall results underlined. Ablations of LELA in gray (see Section~\ref{sec:ablation}).
    }
  \small
  \setlength{\tabcolsep}{3pt}
  \begin{tabular}{lccccc}
    \toprule
    \textbf{Method} & \textbf{Realms} & \textbf{Lego} & \textbf{Star Trek} & \textbf{YuGiOh} & \textbf{Macro Acc.} \\
    \midrule\\[-1mm]
    \multicolumn{6}{l}{\textbf{Zero-shot}} \\
    \midrule
    Baseline**~\cite{logeswaranZeroShotEntityLinking2019} & - & - & - & - & 77.05 \\
    BLINK*~\cite{wuScalableZeroshotEntity2020} & 86.80 & 76.39 & 77.95 & 67.46 & 77.15 \\
    E-repeat**~\cite{yaoZeroshotEntityLinking2020} & - & - & - & - & 79.64 \\
    Uni-MPR**~\cite{tangBidirectionalMultiparagraphReading2021} & 87.25 & 78.57 & 80.56 & 67.31 & 78.42 \\
    Bi-MPR*~\cite{tangBidirectionalMultiparagraphReading2021} & 89.60 & 80.50 & 81.04 & 68.74 & 79.97 \\
    ReS*~\cite{xuReadandSelectFrameworkZeroshot2023} & 88.10 & 78.44 & 81.69 & 75.84 & 81.02 \\
    GenDecider*~\cite{zhouGenDeciderIntegratingNone2024} & \underline{\textbf{91.10}} & \textbf{81.42} & \underline{\textbf{82.48}} & \textbf{75.99} & \textbf{82.75} \\[3mm]
    
    \multicolumn{6}{l}{\textbf{True Zero-shot}} \\
    \midrule
    BLINK***~\cite{wuScalableZeroshotEntity2020} & 84.11 & 74.45 & 72.22 & 66.30 & 74.27 \\
    ExtEnD*~\cite{barbaExtEnDExtractiveEntity2022} & 79.62 & 65.20 & 73.21 & 60.01 & 69.51 \\
    GENRE**~\cite{decaoAutoregressiveEntityRetrieval2021} & 55.20 & 42.71 & 55.76 & 34.68 & 47.09 \\
    OneNet~\cite{liuOneNetFineTuningFree2024} & 55.80 & 53.80 & 53.90 & 40.80 & 51.08 \\
    LLM-SFT (baseline) & 85.59 & 75.03 & 73.93 & 73.70 & 77.06 \\
    \textbf{LELA (ours)} \\
    \textbf{w/ Mag.-Sm.-2509} & \textbf{88.69} & \underline{\textbf{83.04}} & \textbf{79.64} & \underline{\textbf{81.05}} & \underline{\textbf{83.11}} \\
    \rowcolor{lightgray} w/ Qwen3-30B-A3B & 85.69 & 79.14 & 72.68 & 69.65 & 76.79 \\
    \rowcolor{lightgray} w/ Qwen3-4B & 80.08 & 67.93 & 65.28 & 58.69 & 68.00 \\
    \rowcolor{lightgray} w/o LLM selection & 31.23 & 22.10 & 30.56 & 24.70 & 27.15 \\
    \bottomrule
  \end{tabular}%
  \label{tab:zeshel_results}
  \vspace{-\baselineskip}
\end{table}

\subsection{Domain-specific and \textit{None} Scenarios }
\label{sec:esco}

\subsubsection{Benchmark.} The ESCO benchmark~\cite{levrangESCOBoostingJob2014} is designed for the task of mapping occupational skills to the ESCO taxonomy, evaluated on the manually annotated dataset from
\cite{decorteDesignNegativeSampling2022}. Some mentions are mapped to the special entity \texttt{UNK}, because they have no counterpart in the taxonomy (813 out of the 1824 test mentions). 

\subsubsection{LELA Configuration.} We generate our own candidates either with BM25 (retrieving 200 candidates per mention) or with dense retrieval (retrieving 64 candidates per mention). BM25 yields a recall of 80.8\%, and dense retrieval yields 95.6\%.
We use the provided entity descriptions. 

\subsubsection{Results.} Table~\ref{tab:esco_results} shows that, for the accuracy on actual entities, LELA performs on par with the classical approaches.
In the overall accuracy, the baseline that always predicts \texttt{UNK} has the highest accuracy, which is due to the high ratio of \texttt{UNK} mentions. LELA still outperforms the other true zero-shot competitors, as well as the classical approaches, even if the difference in accuracy is not significant.
Dense retrieval of candidates allows for higher non-\texttt{UNK} accuracy across the models, while BM25 results in higher \texttt{UNK} accuracy. 

\newcommand{\conf}[1]{\textcolor{gray}{\fontsize{6}{6}\selectfont$\pm$#1}}

\begin{table}
  \centering
  \caption{Accuracy on the ESCO entity linking benchmark. Accuracy and
      \texttt{UNK} accuracy from~\cite{zhangEntityLinkingJob2024} (Non-\texttt{UNK}
      accuracy is derived), best per category in bold, globally best underlined. We report 95\% confidence intervals to account for the smaller test set size. Ablations of LELA in gray (see Section~\ref{sec:ablation}).}
  \small
  \begin{tabular}{lccc}
    \toprule
    \textbf{Method$~~~~~~~~~~~~~~~~~~~~~~~$Accuracy:} & \textbf{overall} & \textbf{UNK} & \textbf{Non-UNK} \\
    \midrule
    Predict-\texttt{UNK} baseline & \underline{\textbf{44.57}} & \underline{\textbf{100.0}} & \phantom{0}0.0 \\
    Random baseline & 0.22 & - & - \\
    TF-IDF baseline & 2.25 & - & - \\[3mm]
    
    \multicolumn{4}{l}{\textbf{Classical}} \\
    \midrule
    BLINK Wiki+ESCO~\cite{wuScalableZeroshotEntity2020} 
    & \textbf{23.55}\conf{1.95} 
    & \phantom{0}1.38\conf{0.81} 
    & \textbf{41.38}\conf{3.03} \\
    GENRE Wiki+ESCO~\cite{decaoAutoregressiveEntityRetrieval2021} 
    & 11.48\conf{1.46} 
    & \textbf{\phantom{0}1.65}\conf{0.88} 
    & 19.39\conf{2.43} \\[3mm]
    
    \multicolumn{4}{l}{\textbf{True Zero-shot}} \\
    \midrule
    BLINK Wiki~\cite{wuScalableZeroshotEntity2020} 
    & 23.30\conf{1.94} & - & - \\
    GENRE Wiki~\cite{decaoAutoregressiveEntityRetrieval2021} 
    & \phantom{0}6.91\conf{1.16} & - & - \\
    LLM-SFT (baseline)
        & 25.11\conf{2.01} 
        & \phantom{0}5.66\conf{1.59} 
        & 40.75\conf{3.03} \\
    \textbf{LELA (ours)} \\
    \textbf{w/ Mag.-Sm.-2509+Dense  Retrieval}
        & \textbf{26.75}\conf{2.03} 
        & 5.41\conf{1.56} 
        & \underline{\textbf{43.92}}\conf{3.06} \\
    \textbf{w/ Mag.-Sm.-2509+BM25}
        & 26.21\conf{2.02} 
        & \textbf{7.26}\conf{1.79} 
        & 41.44\conf{3.04} \\
    \rowcolor{lightgray} w/ Qwen3-30B-A3B+Dense Retrieval
        & 36.62\conf{2.21}
        & 28.04\conf{3.09}
        & {43.52}\conf{3.06} \\
    \rowcolor{lightgray} w/ Qwen3-30B-A3B+BM25 
        & 39.36\conf{2.24} 
        & 39.11\conf{3.36} 
        & 39.56\conf{3.02} \\
    \rowcolor{lightgray} w/ Qwen3-4B+Dense Retrieval
        & 26.59\conf{2.03} 
        & \phantom{0}8.36\conf{1.91} 
        & 41.25\conf{3.04} \\
    \rowcolor{lightgray} w/ Qwen3-4B+BM25 
        & 29.44\conf{2.09} 
        & 14.88\conf{2.45} 
        & 41.15\conf{3.04} \\
    \rowcolor{lightgray} w/o LLM selection+Dense Retrieval
        & - & - & 35.71\conf{2.96} \\
    \rowcolor{lightgray} w/o LLM selection+BM25 
        & - & - & 34.92\conf{2.94} \\
    \bottomrule
  \end{tabular}%
  
  \label{tab:esco_results}
  \vspace{-\baselineskip}
\end{table}

\subsection{Scenario of Acronym Disambiguation} 

\subsubsection{Benchmark.} GLADIS~\cite{chenGLADISGeneralLarge2023} is a benchmark for the task of
mapping acronyms to their inferred long form, spanning the general, scientific and biomedical domain, with, respectively, 7125, 14066 and 3149 test mentions. The average number of candidates per acronym is 200. 

\subsubsection{LELA Configuration.} We use the short-form to long-form mapping for candidate generation, and use the long-forms as entity descriptions. 

\subsubsection{Competitors.} EL approaches have not been applied to the acronym disambiguation task. Rather, the original paper proposed AcroBERT, which can run both in a classical and in a zero-shot fashion. Since it outperformed alternative approaches by a wide margin, we compare only to it and the baselines in these categories. For the true zero shot group, we run GENRE~\cite{decaoAutoregressiveEntityRetrieval2021} and the BLINK cross-encoder~\cite{wuScalableZeroshotEntity2020}. 

\subsubsection{Results.} Table~\ref{tab:gladis_results} shows that LELA outperforms all true zero-shot competitors by a huge margin of up to 63 percentage points. It also outperforms all other competitors, whether or not they were fine-tuned on the training data, on all domains except the biomedical one. LELA improves the best-reported macro-accuracy by 4.6 percentage points, and the scientific domain accuracy by 19.3 percentage points.

\begin{table}[t]
  \centering
  \caption{Accuracy on the GLADIS benchmark. Best in each group in bold (AcroBERT was previously shown to outperform classical and fine-tuned methods), globally best underlined. Ablations of LELA in gray (see Section~\ref{sec:ablation}).
  }
  \small
  \begin{tabular}{lcccc}
    \toprule
    \textbf{Method} & \textbf{General} & \textbf{Scientific} & \textbf{Biomedical} & \textbf{Macro Acc.} \\
    \midrule
    BM25 baseline~\cite{chenGLADISGeneralLarge2023} & 25.8 & 10.7 & 14.3 & 16.9 \\
    Popularity baseline~\cite{chenGLADISGeneralLarge2023} & 53.2 & 12.3 & 54.0 & 39.8 \\[3mm]
    
    \multicolumn{5}{l}{\textbf{Classical}} \\
    \midrule
    AcroBERT (fine-tuned)~\cite{chenGLADISGeneralLarge2023} & \textbf{76.1} & \textbf{27.6} & \underline{\textbf{69.2}} & \textbf{57.7} \\[3mm]
    
    \multicolumn{5}{l}{\textbf{Zero-shot}} \\
    \midrule
    AcroBERT~\cite{chenGLADISGeneralLarge2023} & \textbf{72.0} & \textbf{27.4} & \textbf{61.4} & \textbf{53.6} \\[3mm]
    
    \multicolumn{5}{l}{\textbf{True Zero-shot}} \\
    \midrule
    GENRE~\cite{decaoAutoregressiveEntityRetrieval2021} (our impl.) & 63.9 & \phantom{0}3.9 & \phantom{0}7.1 & 25.0 \\
    BLINK~\cite{wuScalableZeroshotEntity2020} (our impl.) & 15.8 & \phantom{0}1.5 & \phantom{0}1.8 & \phantom{0}6.4 \\
    LLM-SFT (baseline)  & 59.6 & 17.4 & 31.4 & 36.1 \\
    \textbf{LELA (ours)} \\
    \textbf{w/ Magistral-Small-2509} & \underline{\textbf{78.9}} & \underline{\textbf{46.9}} & \textbf{61.0} & \underline{\textbf{62.3}} \\
    \rowcolor{lightgray} w/ Qwen3-30B-A3B & 80.1 & {45.5} & 61.5 & 62.4 \\
    \rowcolor{lightgray} w/ Qwen3-4B & 73.8 & 37.7 & 54.3 & 55.3 \\
    \rowcolor{lightgray} w/o LLM selection & 44.2 & 22.7 & 22.6 & 29.8 \\
    \bottomrule
  \end{tabular}%

  \label{tab:gladis_results}
  \vspace{-\baselineskip}
\end{table}

\subsection{Wikipedia-based Scenario}

\subsubsection{Benchmark.} ZELDA \cite{milichZELDAComprehensiveBenchmark2023}
groups several established Wikipedia-based classical EL benchmarks.
It aims at allowing for a fair comparison of supervised methods by mapping all mentions to the same Wikidata KB and by providing a common
Wikipedia-based train set and candidate dictionary to be used across benchmarks and methods. We also run LELA on the WikilinksNED Unseen Mentions~\cite{onoeFineGrainedEntityTyping2020} benchmark and report the results in Appendix~\ref{sec:unseen_mentions}.

\subsubsection{LELA Configuration.} We use the provided candidates and entity descriptions.

\subsubsection{Competitors.}
We compare LELA to all entity linking methods that have been trained on ZELDA~\cite{ruckerEvaluatingDesignDecisions2025,milichZELDAComprehensiveBenchmark2023}. These supervised approaches benefit from strong alignment with Wikipedia-based training, as all test set entities are present in the ZELDA training data, which favors memorization-based linking. This makes ZELDA a stress-test for LELA. 

\subsubsection{Results.}
Table~\ref{tab:zelda_results} shows that the best-performing methods are our LLM-SFT baseline and VERBALIZED$_{\textit{it}}$\footnote{VERBALIZED does not use the provided candidate dictionary.}.
Here, LELA cannot achieve the performance of the classical methods.
However, it remains competitive, performing on par with GENRE, despite being the only method that operates without supervised training signals.

\begin{table}[t]
  \centering
  \caption{Accuracy on the ZELDA benchmark. * means reported in~\cite{milichZELDAComprehensiveBenchmark2023}. Best in each group in bold, globally best underlined. Ablations of LELA in gray (see Section~\ref{sec:ablation}).}
  \setlength{\tabcolsep}{2pt}
  \small

  \begin{tabular}{lcccccccccc}
    \toprule
    & AIDA
    & Tweeki
    & \multicolumn{2}{c}{Reddit}
    & \multicolumn{2}{c}{WNED}
    & \multicolumn{3}{c}{Slinks}
    & Avg \\
    \cmidrule(lr){4-5}\cmidrule(lr){6-7}\cmidrule(lr){8-10}
    & 
    & 
    & posts
    & comm.
    & CWEB
    & Wiki
    & Tail
    & Shad.
    & Top
    & \\
    \midrule
      Upper bound* \cite{milichZELDAComprehensiveBenchmark2023} & 91.1 & 94.0 & 98.4 & 98.3 & 92.4 & 98.8 & 98.8 & 56.7 & 73.1 & 89.1 \\
      MFS baseline* \cite{milichZELDAComprehensiveBenchmark2023} & 63.5 & 72.3 & 83.4 & 81.0 & 61.2 & 65.1 & 99.4 & 14.9 & 41.3 & 64.7 \\[3mm]
      
      \multicolumn{11}{l}{\textbf{Classical (with training data)}} \\
      \midrule    
      $\text{FEVRY}_\text{CL}$*~\cite{fevryEmpiricalEvaluationPretraining2020} & 79.5 & 76.9 & 89.0 & 86.5 & 70.3 & 84.5 & 87.6 & 31.9 & 47.7 & 72.7 \\
      $\text{LUKE}_\text{FT}$*~\cite{yamadaGlobalEntityDisambiguation2022} & 81.2 & 77.9 & 81.5 & 78.5 & 70.3 & 76.5 & 98.0 & 22.5 & 51.8 & 71.0 \\
      $\text{GENRE}_\text{CL}$*~\cite{decaoAutoregressiveEntityRetrieval2021} & 78.6 & 80.1 & 92.8 & 91.5 & \underline{\textbf{73.6}} & 88.4 & \underline{\textbf{99.6}} & 37.3 & 52.8 & 77.2 \\
      FusionED~\cite{wangEntityDisambiguationFusion2024} & 80.1 & 81.4 & \underline{\textbf{93.9}} & \underline{\textbf{92.3}} & \underline{\textbf{73.6}} & 89.0 & 98.3 & 41.5 & 57.9 & 78.7 \\
      $\text{BLINK}_\text{bi}$ \cite{ruckerEvaluatingDesignDecisions2025} & 65.5 & 72.1 & 83.1 & 79.1 & 58.1 & 73.1 & 96.3 & 41.8 & 42.6 & 68.0 \\
      VERB'ZED$_\text{IT}$ \cite{ruckerEvaluatingDesignDecisions2025} & \underline{\textbf{88.2}} & 78.9 & 92.2 & 88.4 & 71.5 & \underline{\textbf{90.8}} & 98.2 & \underline{\textbf{66.3}} & \textbf{65.9} & \underline{\textbf{82.3}} \\
      LLM-SFT (basel.) & 85.7 & \textbf{\underline{83.6}} & 92.6 & 92.2 & 71.9 & 89.8 & 97.2 & 44.5 & 63.9 & 80.2 \\[3mm]
      
      \multicolumn{11}{l}{\textbf{True Zero-Shot (without training data)}} \\
      \midrule    
      \textbf{LELA (ours)} \\
      \textbf{w/ Mag-Sm-2509}  & \textbf{83.3} & \textbf{79.1} & \textbf{80.7} & \textbf{79.5} & \textbf{66.7} & \textbf{87.5} & \textbf{97.2} & \textbf{48.9} & \underline{\textbf{66.7}} & \textbf{76.6} \\
      \rowcolor{lightgray} w/ Qwen3-30B-A3B  & 84.2 & 79.1 & 83.9 & 84.6 & 64.8 & 87.5 & 97.3 & 48.8 & 67.0 & 77.5 \\
      \rowcolor{lightgray} w/ Qwen3-4B  & 84.0 & 78.0 & 84.1 & 82.6 & 66.3 & 85.1 & 97.1 & 48.2 & 65.0 & 76.7 \\
      \rowcolor{lightgray} w/o LLM selection & 76.4 & 75.1 & 79.0 & 79.9 & 65.0 & 76.2 & 97.0 & 43.6 & 60.6 & 72.5 \\
    \bottomrule
  \end{tabular}%
  \vspace{-\baselineskip}
  \label{tab:zelda_results}
\end{table}

\subsection{Ablation Studies and Analyses}
\label{sec:ablation}

\subsubsection{Model Choice.} Tables~\ref{tab:zeshel_results}, \ref{tab:esco_results}, \ref{tab:gladis_results}, and \ref{tab:zelda_results} show that our default model \texttt{Magistral\-Small-2509} stands out on ZESHEL, where differences in performance between the models are the largest. On the other datasets, \texttt{Qwen3-30B-A3B} catches up. It also handles \texttt{None} prediction better. On the more general domains, the smaller \texttt{Qwen3-4B} also performs well. 
Overall, the performance of LELA is robust to the choice of the underlying LLM.
The tables show also that omitting the selection LLM (``w/o LLM selection'') lowers the results drastically on domain-specific datasets, indicating that LELA's joint consideration of candidates is vital. 

\subsubsection{Ablations.} We study several ablations on the ZESHEL benchmark:

\begin{description}
    \item[No Reranker:] We pass all retrieved candidates to the LLM without reranking nor cutoff.
    \item[No Self-Consistency:] We sample only one LLM answer.
    \item[No Reasoning:] We skip thinking token generation and prompt the selection LLM to directly output an answer.
    \item[No Descriptions:] We pass empty entity descriptions to the selection LLM.
\end{description}
Table~\ref{tab:ablation} shows that \textbf{removing the reranker} results in a drop in performance (in addition to a longer runtime and increased cost given the longer contexts). While the drop in performance is significant for \texttt{Qwen3-30B-A3B} and \texttt{Qwen3-4B}, it is negligible for \texttt{Magistral-Small-2509}, which is due to {the fact that this larger model is better at handling a large number of candidates, resulting in a slower drop in selection accuracy as $k$ increases, which can thus be offset by the increased candidate retrieval recall (see Appendix~\ref{sec:recall_acc_tradeoff}).
The \textbf{No Self-Consistency} variant, too, systematically results in a drop in performance (more details in Appendix~\ref{sec:self_consistency}). 
\textbf{Disabling reasoning} results in the next biggest drop in performance across the three models. This shows that simply considering the candidates together is not sufficient and that context-based reasoning is needed.
Finally, the \textbf{No description variant} provides the selection LLM with less precise entities, and it also performs worse in all configurations. Together these results show that all components of LELA contribute to its performance.
\textbf{Examples} of candidate selection outputs for different variants are given in Appendix~\ref{sec:examples}.

\begin{table}
  \centering
  \small
  \caption{Normalized accuracy on the test set of ZESHEL across LELA variants. The results for the \textit{w/o reasoning} variant are reported for the Qwen models, which support switching between reasoning and non-reasoning mode.
  }
  \begin{tabular}{lccccc}
    \toprule
    Method &  Realms & Lego & Star Trek & YuGiOh & Macro Acc. \\
    \midrule
    \textbf{LELA} \\
    w/ Magistral-Small-2509 
      & \textbf{88.69} & 83.04 & 79.64 & 81.05 & \textbf{83.11} \\
    \hspace{3mm} w/o reranker 
      & 87.79 & \textbf{83.45} & \textbf{80.00} & \textbf{81.15} & 83.10 \\
    \hspace{3mm} w/o self-consistency 
      & 81.38 & 76.98 & 73.29 & 74.28 & 76.48 \\
    \hspace{3mm} w/o descriptions 
      & 76.58 & 71.22 & 72.53 & 71.24 & 72.89 \\
    \addlinespace[1mm]
    w/ Qwen3-30B-A3B 
      & \textbf{85.69} & \textbf{79.14} & \textbf{72.68} & \textbf{69.65} & \textbf{76.79} \\
    \hspace{3mm} w/o reranker 
      & 83.58 & 74.31 & 67.07 & 64.44 & 72.35 \\
    \hspace{3mm} w/o self-consistency 
      & 82.08 & 73.48 & 65.39 & 63.18 & 71.03 \\
    \hspace{3mm} w/o descriptions 
      & 71.67 & 62.28 & 65.92 & 60.64 & 65.13 \\
    \hspace{3mm} w/o reasoning 
      & 65.97 & 53.24 & 52.06 & 45.64 & 54.23 \\
    \addlinespace[1mm]
    w/ Qwen3-4B 
      & \textbf{80.08} & \textbf{67.93} & \textbf{65.28} & \textbf{58.69} & \textbf{68.00} \\
    \hspace{3mm} w/o reranker 
      & 75.78 & 60.43 & 61.01 & 51.14 & 62.09 \\
    \hspace{3mm} w/o self-consistency 
      & 77.48 & 62.49 & 60.25 & 53.82 & 63.51 \\
    \hspace{3mm} w/o descriptions 
      & 66.87 & 54.16 & 61.08 & 51.57 & 58.42 \\
    \hspace{3mm} w/o reasoning 
      & 51.55 & 36.79 & 36.45 & 35.75 & 40.14 \\
    \bottomrule
  \end{tabular}%
  
  \label{tab:ablation}
  \vspace{-\baselineskip}
\end{table}

\label{sec:analysis}

\subsubsection{Trading off Recall vs Selection Accuracy.} Lower values of $k$ allow for a high selection accuracy, but may result in the correct entity not being among the candidates, while higher values reduce the selection accuracy but improve the candidate recall. We study the impact of $k$ in Appendix~\ref{sec:recall_acc_tradeoff}: overall accuracy peaks around $k=10$ for all models, with lower values leading to a substantial degradation in recall, and higher values to a lower overall performance due to the drop in selection accuracy. The reranking stage is what makes this cutoff safe: on ZESHEL/Forgotten Realms it preserves 98\% of the recoverable recall while compressing the candidate list by a factor of 6.4, though retention is domain-dependent, dropping to 63.2\% on GLADIS/Scientific (Appendix~\ref{sec:retention_funnel}).

\subsubsection{Self-consistency.} 
We study the impact of self-consistency over different numbers of sampled answers in Appendix~\ref{sec:self_consistency}. We observe that while performance increases with the number of samples, an accuracy plateau is reached quickly across models, suggesting that as few as 3 samples suffice.

\subsubsection{Positional Bias.} LLMs are subject to positional bias~\cite{liuLostMiddleHow2023}. In Appendix~\ref{sec:positional_bias}, we shuffle the list of candidates given to the LLM. Larger models appear robust to this bias, while smaller models are more sensitive to it. Overall, the results justify our approach of sorting the candidates according to their reranker score. 

\subsubsection{Runtime and Scalability.} Table~\ref{tab:runtime_main} reports the efficiency--accuracy trade-off across benchmarks. LELA is one to two orders of magnitude slower than shallow retrieval methods: on GLADIS, BLINK and GENRE process a mention in 0.9 and 4.0 seconds, against 13.9 to 57.2 seconds for LELA depending on the selection LLM. This cost buys a large accuracy gain (from 6.4\% and 25.0\% to 64.5\% macro accuracy), and remains within a respectable envelope for the offline and batch processing that LELA targets. The comparison with the \textit{w/o LLM selection} variant isolates the cost of the selection stage itself: on ZESHEL, it accounts for the difference between 1.3 seconds at 27.15\% accuracy and 7.1 seconds at 78.30\%. The MoE models offer the most compelling speed/performance trade-off: \texttt{gpt-oss-20b} runs 3 to 6 times faster than \texttt{Magistral-Small-2509} per mention, while exceeding its accuracy on three of the four benchmarks, and it also runs faster than the much smaller dense \texttt{Qwen3-4B}. The latter should therefore only be preferred in cases of VRAM limitations. Appendix~\ref{sec:runtime} reports thought token usage, and shows that selection accuracy decreases as the input context length increases, with a model-dependent rate of decline, along with traces of \textit{underthinking} and \textit{overthinking} behaviors in reasoning LLMs~\cite{suUnderthinkingOverthinkingEmpirical2025a}.

\begin{table}
  \centering
  \small
  \caption{Efficiency--accuracy trade-off: average processing time per mention (s) and accuracy, across benchmarks. Metrics are as in Tables~\ref{tab:zeshel_results}--\ref{tab:zelda_results}. $\dagger$ indicates a classical (non-true-zero-shot) configuration. Competitor timings were measured on GLADIS. Thought token usage is reported in Appendix~\ref{sec:runtime}.}
  \setlength{\tabcolsep}{3.5pt}
  \begin{tabular}{lcccccccc}
    \toprule
    & \multicolumn{2}{c}{\textbf{ZESHEL}} & \multicolumn{2}{c}{\textbf{GLADIS}} & \multicolumn{2}{c}{\textbf{Unseen M.}} & \multicolumn{2}{c}{\textbf{ESCO}} \\
    \cmidrule(lr){2-3}\cmidrule(lr){4-5}\cmidrule(lr){6-7}\cmidrule(lr){8-9}
    \textbf{Method} & s & Acc. & s & Acc. & s & Acc. & s & Acc. \\
    \midrule
    BLINK~\cite{wuScalableZeroshotEntity2020} & - & 74.27 & \phantom{0}0.9 & \phantom{0}6.4 & - & 75.2$^\dagger$ & - & 23.30 \\
    GENRE~\cite{decaoAutoregressiveEntityRetrieval2021} & - & 47.09 & \phantom{0}4.0 & 25.0 & - & 63.5$^\dagger$ & - & \phantom{0}6.91 \\
    LLM-SFT (baseline) & - & 77.06 & \phantom{0}1.2 & 36.1 & - & 69.5$^\dagger$ & - & 25.11 \\
    \midrule
    \textbf{LELA (ours)} \\
    \textbf{w/ Magistral-Small-2509} & 44.4 & 83.11 & 57.2 & 62.3 & 35.8 & 68.0 & 54.6 & 26.21 \\
    w/ Qwen3-30B-A3B & 26.0 & 76.79 & 19.2 & 62.4 & 10.0 & 68.7 & 21.7 & 39.36 \\
    w/ Qwen3-4B & 31.1 & 68.00 & 14.7 & 55.3 & 19.6 & 67.0 & 27.2 & 29.44 \\
    w/ gpt-oss-20b & \phantom{0}7.1 & 78.30 & 13.9 & 64.5 & \phantom{0}8.2 & 69.4 & 16.6 & 31.30 \\
    w/o LLM selection & \phantom{0}1.3 & 27.15 & \phantom{0}2.3 & 29.8 & \phantom{0}1.3 & 61.9 & \phantom{0}0.5 & - \\
    \bottomrule
  \end{tabular}
  \label{tab:runtime_main}
  \vspace{-\baselineskip}
\end{table}

\subsubsection{Reranker Performance.} We compare the performance of different rerankers  in Appendix~\ref{sec:reranker}. Other models achieve a recall@k similar to the one of our default model \texttt{Qwen3-Reranker-4B}, showing that LELA is not tied to a specific reranker.

\subsubsection{Discussion.}
Our results suggest that LLM-based candidate reranking and selection can effectively compensate for the absence of labeled training data, and even outperform fine-tuned approaches. Our ablation studies suggest that these gains stem primarily from the ability of LLMs to jointly reason over multiple candidates rather than from candidate ranking alone, while our baseline comparisons highlight the superiority of reasoning-based selection over standard domain-transfer for true zero-shot EL.
Furthermore, the fact that candidate ranking alone does not perform well also mitigates concerns regarding data contamination: if the models were simply recalling memorized gold labels from pre-training, the reranking alone would nearly always find the best candidate -- which it does not.
These findings naturally raise two follow-up questions: 

\subsubsection{How much in-domain annotation does it take to match LELA?}
$~$\\
\noindent While we frame LELA as an alternative to annotation in cold-start settings, it is natural to ask at what point in-domain supervision overtakes inference-time scaling. We conduct an experiment on the Star Trek domain of ZESHEL (the largest test domain), starting from our LLM-SFT baseline pre-trained on Wikipedia mentions, and fine-tuning it on increasing amounts of in-domain data.
Figure~\ref{fig:sft_exp} compares this against LELA. The pre-trained baseline already outperforms the \texttt{Qwen3-4B} variant, and 16 in-domain samples allow it to catch up with the \texttt{Qwen3-30B-A3B} variant. However, more than 64 mentions are needed to match the \texttt{gpt-oss-20b} variant, and over 400 mentions to catch up with \texttt{Magistral-Small-2509}, with no further improvements during the following steps. This makes the trade-off concrete: when sufficient labeled data can be afforded, in-domain fine-tuning is highly relevant; in cold-start settings, LELA delivers performance equivalent to hundreds of annotated mentions through inference-time compute. Full experimental details are reported in Appendix~\ref{sec:sft_experiment}.

\begin{figure}
    \centering
    \vspace{-\baselineskip}
    \includegraphics[width=0.49\textwidth]{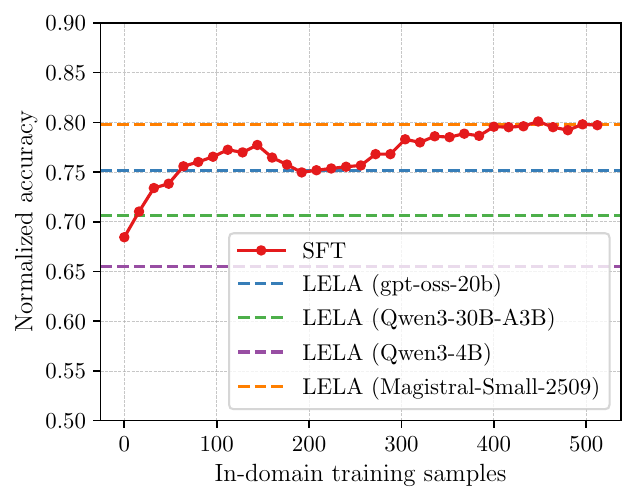}
    \caption{LLM-SFT vs LELA accuracy on the Star Trek domain of ZESHEL as a function of the number of in-domain training mentions.}
    \label{fig:sft_exp}
    \vspace{-\baselineskip}
\end{figure}

\subsubsection{Can we retain LELA's gains at lower inference cost?\\}
$~$\\
\begin{wraptable}{r}{0.45\textwidth}
        \vspace{-2\baselineskip}
        \caption{Macro-averaged normalized accuracy for distilled student models on ZESHEL.}
        \small
        \begin{tabular}{lc}
        \toprule
        Model & Accuracy \\
        \midrule
        \multicolumn{2}{l}{\emph{Reference (no distillation)}} \\
        \texttt{Qwen3-4B} & 68.00 \\
        \texttt{gpt-oss-20b} (teacher) & 78.30 \\
        \midrule
        \multicolumn{2}{l}{\emph{Distilled students}} \\
        \texttt{Qwen3-0.6B} & 72.78 \\
        \texttt{Qwen3-1.7B} & 75.15 \\
        \texttt{Qwen3-4B} & 78.61 \\
        \bottomrule
        \end{tabular}
        \label{tab:distill}
        \vspace{-\baselineskip}
\end{wraptable}
While our results show that larger models achieve higher accuracy, deploying them at scale remains expensive. We thus investigate whether LELA's reasoning-based selection capability can be transferred to smaller students through 
distillation~\cite{hintonDistillingKnowledgeNeural2015,kimSequenceLevelKnowledgeDistillation2016} while also preserving the true zero-shot constraint on the test domain. 
We sample 10{,}000 Wikipedia mentions from the ZELDA training split, generate teacher completions using \texttt{gpt-oss-20b} with the top-10 reranker 
candidates, filter for correctness, and use the reasoning traces to fine-tune three Qwen3 students of decreasing size. 
Table~\ref{tab:distill} reports the results: the distilled \texttt{Qwen3-4B} student matches the teacher, gaining 
10.6 percentage points over the base model, while the smaller \texttt{Qwen3-1.7B} and \texttt{Qwen3-0.6B} students retain 96\% and 93\% of the teacher's accuracy respectively. Even the 0.6B student outperforms several supervised baselines from Table~\ref{tab:zeshel_results}.
This shows that LELA's reasoning-based selection capability can be distilled into smaller models with minimal loss, transferring across domains, providing a practical path to deploying LELA at significantly reduced inference cost. Distillation hyper-parameters are reported in Appendix~\ref{sec:distillation}.

\section{Conclusion}

We have proposed LELA, a simple and modular \textit{coarse-to-fine} approach to entity linking.
LELA is \emph{true zero-shot}, i.e., it can be applied to a given knowledge base and domain without specific annotated data or fine-tuning.
LELA outperforms other true zero-shot approaches by a large margin, and performs on par, and often beats, fine-tuned and classical approaches, even though these require training data. 
Our work points to a fundamental shift in how entity linking can be approached in resource-constrained environments: While traditional methods optimize for inference efficiency, front-loading costs onto human annotation, LELA demonstrates that allocating computational budget to test-time reasoning can effectively substitute for domain-specific supervision. This suggests that the ability for entity disambiguation is a general capability of reasoning models that transfers across settings, unlike the distributional patterns learned during fine-tuning. Consequently, the usual barrier to entry for entity linking in specialized domains (viz. the availability of labeled data) could be mitigated with additional inference compute.

This shift comes with a concrete cost, and our experiments delimit when it is worth paying. LELA is one to two orders of magnitude slower per mention than shallow retrieval methods, and reaches the accuracy that in-domain fine-tuning would require hundreds of annotated mentions to match. It is therefore the appropriate choice when annotation is the scarcer resource: cold-start deployments, proprietary or rapidly evolving knowledge bases, specialized domains where labeling requires expert time, and offline or batch processing where per-mention latency is not binding. Conversely, when a labeled corpus can be afforded and mentions must be linked at high throughput, a fine-tuned model remains preferable. Between these two regimes, our distillation experiments offer a middle ground: because the students are trained on out-of-domain Wikipedia mentions only, distillation reduces inference cost while leaving the true zero-shot guarantee on the target domain intact.

We thus hope that LELA opens the door for new fine-tuning-free EL approaches. In particular, we encourage LLMs to be included as baselines in entity linking studies. 
Future work could extend LELA to end-to-end entity linking, where mentions have to be detected as well.

\begin{credits}
\subsubsection{\ackname} The work was partially supported by Agence de l’Innovation
de Défense – AID - via Centre Interdisciplinaire d’Etudes
pour la Défense et la Sécurité – CIEDS - (project 2024 - KB-LM).

\subsubsection{Supplemental Material Statement.}

Our code is available at our GitHub Repository \url{https://github.com/dig-team/LELA}.

\subsubsection{Declaration of use of Generative AI.}

Generative AI assistants (GitHub Copilot, Claude) were used to help with parts of the code implementation.

\end{credits}
%
%
%
\bibliographystyle{splncs04}
\bibliography{mybibliography}

\appendix

\section{Prompts}
\label{sec:prompts}

The reranker task, retriever task and selection LLM prompt are given in Figure~\ref{fig:all_prompts}.

\begin{figure}[h]
\centering
\begin{subfigure}{0.9\linewidth}
\caption{Reranker task}
\raggedright
\begin{framed}
\footnotesize
Given a text with a marked mention enclosed in square brackets, retrieve relevant entities that the mention refers to.
\end{framed}
\label{fig:reranker_prompt}
\end{subfigure}
\begin{subfigure}{0.9\linewidth}
\caption{Retriever task}
\raggedright
\begin{framed}
\footnotesize
Given an ambiguous mention, retrieve relevant entities that the mention refers to.
\end{framed}
\label{fig:retriever_prompt}
\end{subfigure}
\begin{subfigure}{0.9\linewidth}
\caption{Selection LLM prompt}
\raggedright
\begin{framed}
\footnotesize
You are an expert designed to disambiguate entities in text, taking into account the overall context and a list of entity candidates. You are provided with an input text that includes a full contextual narrative, a marked mention enclosed in square brackets, and a list of candidates, each preceded by an index number.\\[1ex]
Your task is to determine the most appropriate entity from the candidates based on the context and candidate entity descriptions.
Please show your choice with only the index, e.g., ``answer: 3''.
\end{framed}
\label{fig:prompt}
\end{subfigure}
\caption{LELA's prompts}
\label{fig:all_prompts}
\vspace{-\baselineskip}
\end{figure}

\section{The LLM-SFT Baseline}
\label{sec:baseline}

The LLM-SFT baseline was implemented for cross-benchmark comparison of LELA to another LLM-based EL system. It uses \texttt{Qwen3-4B} as the backbone model and the same prompt as LELA (Figure~\ref{fig:prompt}, Section~\ref{sec:selection}). It was trained with LoRA~\cite{huLoRALowRankAdaptation2021} (with $r = 16$ and $\alpha = 32$) using the TRL\footnote{\url{https://huggingface.co/docs/trl/index}} library, for two epochs and a maximum context length of 8192 tokens, on 20{,}000 Wikipedia mentions sampled from the ZELDA training split. For training, we use the ZELDA prior-based candidates and filter out the mentions with fewer than 10 candidates and with a context document exceeding 5120 tokens. 
We set the learning rate to $2 \times 10^{-4}$ and use an effective batch size of 16. The training was conducted on a single L40S 48GB GPU.
Checkpoint selection was guided by accuracy on a 1.5\% held-out subset of the training data.

\section{Category-specific Results on ZESHEL}
\label{sec:category_specific}

In ZESHEL, mentions are split into four categories, based on token
overlap with the entity name:

\begin{description}
\item [High Overlap (HO):] the entity name is identical to the mention text
    \item[Multiple Categories (MC):] the entity name is the mention text followed by a
        disambiguation phrase (e.g. mention string: ``Batman'', title: ``Batman
        (Lego)'')
    \item[Ambiguous Substrings (AS):] the mention text is a substring of the entity name
    \item[Low Overlap (LO):] all other cases
\end{description}

\noindent These categories respectively make up 5\%, 28\%, 8\% and 59\% of the mentions
in the dataset. 
The category-specific normalized accuracy in Table~\ref{tab:zeshel_category_results} shows that using \texttt{Magistral-Small-2509}, LELA ranks first on two of the four categories, and shines in the MC
case. Our LLM-SFT baseline ranks first on the two remaining cases, where LELA is a close second. We conjecture that high performance on the AS case can be explained by the high exposure of LLMs to such cases during pre-training, which occur in all kinds of text corpora, while LELA's performance on the MC case can be attributed to the need for understanding fine-grained distinctions between particularly similar candidates, and which can benefit from reasoning. Regarding the LO case, with lower relative improvement, and where the LLM-SFT baseline underperforms, there is minimal lexical overlap which could lead to (1) embedding-based approaches also being well-suited since they are optimized for semantic similarity, (2) the LLM-SFT baseline suffering from the domain shift for those non-trivial mentions. 

\begin{table}
  \centering
  \small
  \caption{Category-specific macro-averaged normalized accuracy on the test set
  of ZESHEL, results for the other methods as reported in~\cite{xuReadandSelectFrameworkZeroshot2023}.}
  \begin{tabular}{lcccc}
    \toprule
    Method & HO & MC & AS & LO \\
    \midrule
    Baseline & 87.64 & 77.27 & 75.89 & 71.46 \\
    BLINK & 94.30 & 75.40 & 79.95 & 73.50 \\
    Uni-MPR & 91.43 & 79.07 & 75.60 & 73.53 \\
    Bi-MPR & 92.84 & 81.93 & 77.37 & 73.88 \\
    ReS & 94.42 & 81.29 & 77.80 & 76.51 \\
    LLM-SFT & \textbf{96.22} & 78.40 & \textbf{85.15} & 71.01 \\
    \midrule
    \textbf{LELA (ours)} \\
    w/ Magistral-Small-2509 & 95.63 & \textbf{85.19} & 84.95 & \textbf{77.99} \\
    \rowcolor{lightgray} w/ Qwen3-30B-A3B & 88.39 & 79.16 & 76.88 & 68.31 \\
    \rowcolor{lightgray} w/ Qwen3-4B & 79.03 & 70.83 & 70.68 & 60.92 \\
    \bottomrule
  \end{tabular}
  \label{tab:zeshel_category_results}
  \vspace{-\baselineskip}
\end{table}

\section{Unseen Mentions and Short Context Scenario}
\label{sec:unseen_mentions}

\subsubsection{Benchmark.} The WikilinksNED Unseen-Mentions dataset was proposed in~\cite{onoeFineGrainedEntityTyping2020}. It is derived from the WikilinksNED dataset~\cite{eshelNamedEntityDisambiguation2017}, ensuring that the mention-entity pairs are unseen at test time. The examples in the dataset consist of single sentences, resulting in highly limited context. The target knowledge base is Wikipedia.

\subsubsection{LELA configuration.} We retrieve 64 candidates per mention through dense retrieval over Wikipedia entities, using the ELEVANT EL evaluation tool~\cite{bastELEVANTFullyAutomatic2022} Wikipedia mappings\footnote{\url{https://github.com/ad-freiburg/elevant/wiki/Mapping-Files\#wikipedia-mappings}}, which achieves a recall of 83\%. We also experiment with prior-based candidates using the candidate list from~\cite{decaoAutoregressiveEntityRetrieval2021}, which yields a recall of 84\%.

\subsubsection{Results.} Results reported in Table~\ref{tab:ned_results} show that BLINK leads on the benchmark. Regarding LELA, using the prior-based candidates leads to the best results. In particular, these candidates allow LELA to outperform the true zero-shot ET4EL, as well as the classical GENRE and ReFinED methods, which also use prior-based candidates. In fact, for all three models considered, using dense retrieval systematically underperforms on this benchmark. 
Therefore, we conducted an error analysis to analyze LELA's behavior when using dense retrieval in this setting. Four out of the ten analyzed errors come from candidate retrieval, and another four come from imprecise ground truth annotations, where we believe that our approach's predicted Wikipedia entity could also be considered as correct. This is caused by the large number of highly similar entities in Wikipedia.
Only the two remaining cases correspond to the reranker losing the correct candidate and a wrong LLM final answer selection.

\begin{table}
  \centering
  \caption{Accuracy on the WikilinksNED Unseen Mentions benchmark, best per category in bold, globally best underlined.}
  \small
  \begin{tabular}{lcc}
    \toprule
    \textbf{Method} & \textbf{Type} & \textbf{Acc.} \\
    \midrule
    \parbox{4cm}{Most Frequent Entity Baseline\\ \cite{onoeFineGrainedEntityTyping2020}} & baseline & \textbf{54.1} \\
    \midrule
    BLINK~\cite{wuScalableZeroshotEntity2020} & & \\
    \hspace{3mm} Wikipedia & classical & 75.2 \\
    \hspace{3mm} Training set & classical & 74.7 \\
    \hspace{3mm} Training set + Wikipedia & classical & \textbf{\underline{76.8}} \\
    \midrule
    GENRE~\cite{decaoAutoregressiveEntityRetrieval2021} & classical & 63.5 \\
    ReFinED~\cite{ayoolaReFinEDEfficientZeroshotcapable2022} & classical & 66.5 \\
    LLM-SFT & classical & 69.5 \\
    \midrule 
    ET4EL~\cite{onoeFineGrainedEntityTyping2020} & true zero-shot & 62.2 \\
    \midrule
    \textbf{LELA (ours)} & true zero-shot \\
    \textbf{w/ Magistral-Small-2509} \\
    \hspace{3mm} Dictionary & & \textbf{68.0} \\
    \hspace{3mm} Dense Retrieval & & 62.8 \\
    \rowcolor{lightgray} w/ Qwen3-30B-A3B & & \\
    \rowcolor{lightgray} \hspace{3mm} Dictionary & & 68.7 \\
    \rowcolor{lightgray} \hspace{3mm} Dense Retrieval & & 61.9 \\
    \rowcolor{lightgray} w/ Qwen3-4B & & \\
    \rowcolor{lightgray} \hspace{3mm} Dictionary & & 67.0 \\
    \rowcolor{lightgray} \hspace{3mm} Dense Retrieval & & 59.0 \\
    \rowcolor{lightgray} w/o LLM selection & & \\
    \rowcolor{lightgray} \hspace{3mm} Dictionary & & 61.9 \\
    \rowcolor{lightgray} \hspace{3mm} Dense Retrieval & & 54.5 \\
    \bottomrule
  \end{tabular}
  \label{tab:ned_results}
\end{table}

\section{The Recall \emph{vs} Selection Accuracy Trade-off}
\label{sec:recall_acc_tradeoff}

The performance of LELA is a product of the candidate retrieval recall (\emph{is the correct answer within the retrieved candidates?}) and the candidate selection accuracy (\emph{does the LLM select the correct answer?}). Regarding the first retrieval stage, the number of retrieved candidates should be exclusively limited by computational cost constraints, given the finer subsequent stages and that the recall can only improve with the number of retrieved candidates. However, for the next two stages, the value of $k$ results in a trade-off between recall and selection accuracy, as shown in Figure~\ref{fig:acc_recall}: Lower values of $k$ allow for a high selection accuracy, but may result in the correct entity not being among the candidates. Higher values of $k$ reduce the selection accuracy but improve the candidate recall. The impact of $k$ on these two components depends on the reranker and selection LLM in use. For instance, the selection accuracy experiences a steeper drop when $k$ increases for \texttt{Qwen3-30B-A3B} than \texttt{Magistral-Small-2509}. Reranker-specific results are given in Appendix~\ref{sec:reranker}. Figure~\ref{fig:selection_example} shows an example of selection output for $k=10$ and $k=64$.

\begin{figure}
    \centering
    \begin{subfigure}[t]{0.49\linewidth}
        \centering
        \includegraphics[width=\linewidth]{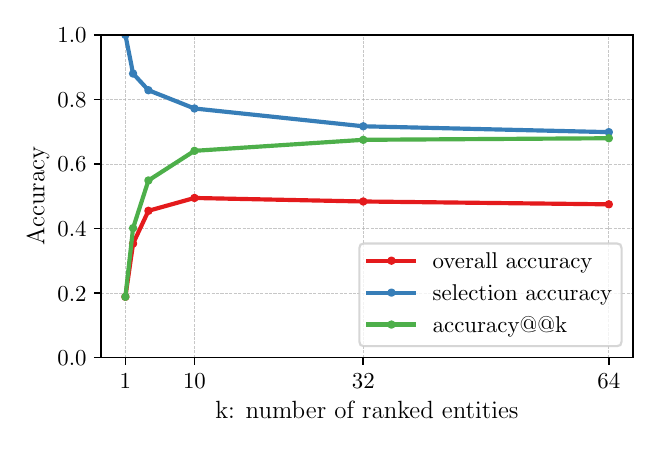}
        \caption{Overall (micro-averaged) accuracy, selection accuracy and accuracy@k w.r.t. k for \texttt{Qwen3-30B-A3B}}
        \label{fig:acc_recall_model1}
    \end{subfigure}
    \hfill
    \begin{subfigure}[t]{0.49\linewidth}
        \centering
        \includegraphics[width=\linewidth]{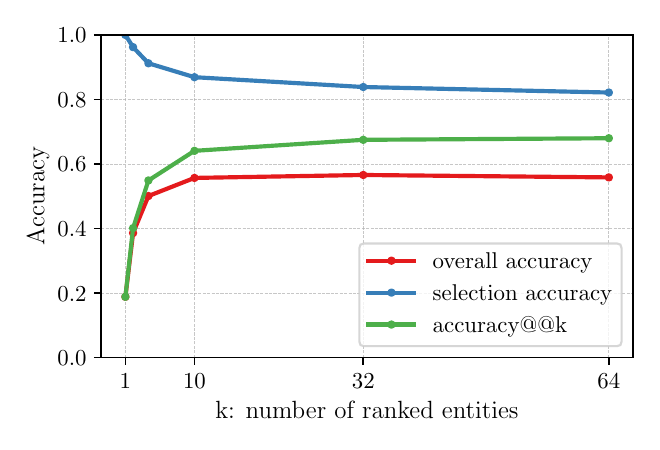}
        \caption{Overall (micro-averaged) accuracy, selection accuracy and accuracy@k w.r.t. k for \texttt{Magistral-Small-2509}}
        \label{fig:acc_recall_model2}
    \end{subfigure}

    \caption{Comparison of (micro-averaged) accuracy, selection accuracy and accuracy@k on ZESHEL.}
    \label{fig:acc_recall}
    \vspace{-\baselineskip}
\end{figure}

\section{Self-consistency Study}
\label{sec:self_consistency}

As shown by our ablation studies in Section~\ref{sec:ablation}, self-consistency systematically allows for performance improvements over single selection LLM answer sampling.
Figure~\ref{fig:acc_sc} reports the selection accuracy over the number of sampled answers, $k_{sc}$, and shows that a plateau is reached across models for $k_{sc}=3$. The best improvement comes from $k_{sc}=2$. Furthermore, Figure~\ref{fig:invalid_sc} shows how the invalid output rate, which also significantly drops for $k_{sc}=2$. In order to check that the impact of self-consistency is not strictly limited to higher robustness towards invalid model outputs, we show in Figure~\ref{fig:diversity_sc} that for our default $k_{sc} = 10$ choice, the models select the same candidate across all samples in approximately 60\% of the cases, two answers in approximately 25\% of cases and three answers in less than 10\% of the cases.

\begin{figure}
    \centering
    \includegraphics[width=.49\textwidth]{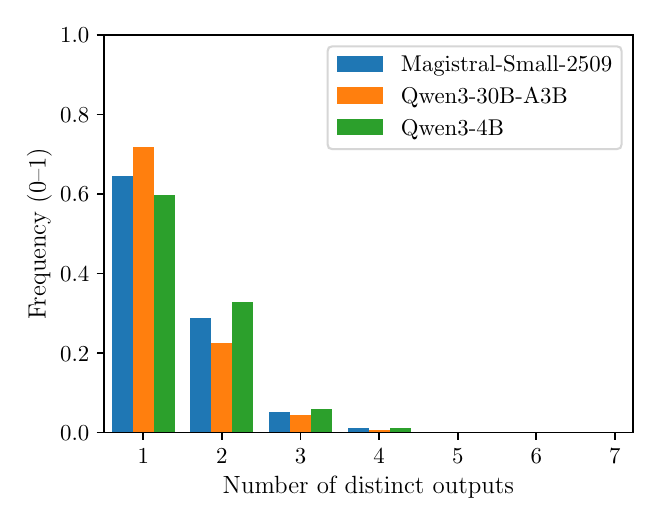}
    \caption{Output diversity histogram, using 10 samples for self consistency, on ZESHEL.}
    \label{fig:diversity_sc}
    \vspace{-\baselineskip}
\end{figure}

\begin{figure}
    \centering
    \begin{subfigure}[b]{0.49\linewidth}
        \centering
        \includegraphics[width=\linewidth]{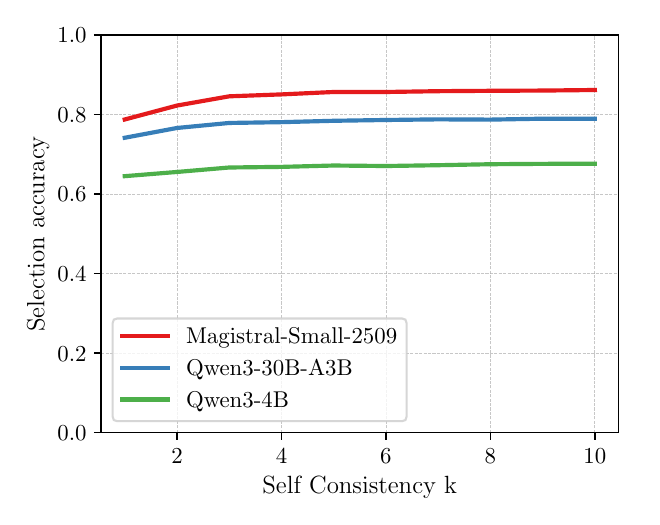}
        \caption{Selection accuracy based on the number of self-consistency samples.}
        \label{fig:acc_sc}
    \end{subfigure}
    \hfill
    \begin{subfigure}[b]{0.49\linewidth}
        \centering
        \includegraphics[width=\linewidth]{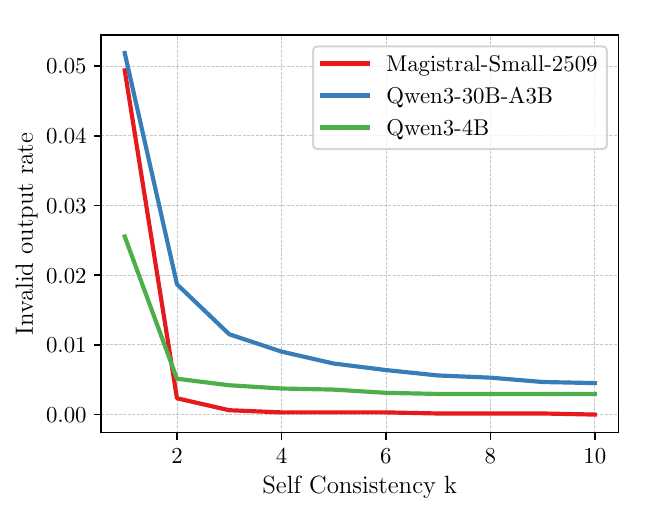}
        \caption{Invalid LLM output rate based on the number of self-consistency samples.}
        \label{fig:invalid_sc}
    \end{subfigure}
    \caption{Effect of the number of self-consistency samples on selection accuracy and invalid output rate on ZESHEL.}
    \label{fig:combined_sc}
    \vspace{-\baselineskip}
\end{figure}

\section{Positional Bias Study}
\label{sec:positional_bias}

To study the impact of the positional bias~\cite{liuLostMiddleHow2023} on the LLM selection accuracy, we report the results on ZESHEL for five different top-k candidate sorting approaches:

\begin{itemize}
    \item \textbf{Reranker:} We rank the top-k candidates according to their reranker-assigned score (the standard LELA approach)
    \item \textbf{First:} The correct answer is moved to the first position.
    \item \textbf{Last:} The correct answer is moved to the last position.
    \item \textbf{Random:} The top-k candidates are shuffled.
    \item \textbf{BM25:} We rank the top-k candidates according to their BM25 score.
\end{itemize}

The results in Table~\ref{tab:positional_bias_zeshel} show varying effects of candidate ordering across models. For \texttt{Magistral-Small-2509}, performance differences between the sorting approaches are minimal, indicating a low sensitivity to candidate position. Among these, the oracle \textit{Answer first} sorting achieves the highest accuracy, closely followed by the \textit{Reranker} sorting.
For \texttt{Qwen3-30B-A3B}, the performance gap between sorting strategies becomes more pronounced. Again, the \textit{Answer first} sorting yields the highest accuracy, with the \textit{Reranker} sorting ranking second. In contrast, \textit{Random} sorting performs worst, trailing the oracle setting by 2.51 percentage points.
This effect further increases for \texttt{Qwen3-4B}, which exhibits the largest performance spread across sorting strategies. For this smaller model, the \textit{Reranker} sorting achieves the highest accuracy, followed by the \textit{Answer first} sorting, while \textit{Random} sorting again results in the lowest performance, 4.84 percentage points behind the best-performing approach.

Overall, these results demonstrate that candidate position has a measurable impact on selection accuracy, with sensitivity increasing as model size decreases. We hypothesize that larger models are better able to compare candidates independently of their order, while smaller models rely more heavily on positional cues. The intermediate behavior of the 30B MoE \texttt{Qwen3-30B-A3B} model is consistent with this explanation. Importantly, the proposed \textit{Reranker} sorting strategy consistently ranks first among the non-oracle methods and achieves the best performance for the smallest model, justifying its use in LELA. In contrast, \textit{BM25} sorting underperforms relative to the reranker-based approaches, while \textit{Random} sorting yields the lowest accuracy across all models.

Furthermore, we also report the results on ESCO to study the impact in the case of \textit{None} prediction, considering three different positions for the \texttt{UNK} candidate, among the candidates according to the reranker-assigned scores:

\begin{itemize}
    \item \textbf{\texttt{None} first:} The \texttt{None} option is in first position (the standard LELA approach)
    \item \textbf{\texttt{None} last:} The \texttt{None} option is in last position
    \item \textbf{Random:} The \texttt{None} option is in a random position
\end{itemize}

\noindent The results are reported in Table~\ref{tab:positional_bias_mhercl}. Across all models, changing the position of the \textit{None} candidate produces only marginal differences in overall accuracy, indicating no statistically significant effect. We observe a consistent trend in which placing \textit{None} first slightly improves \texttt{UNK} accuracy at the cost of marginally reduced non-\texttt{UNK}. Overall, model performance appears robust to the ordering of the None candidate.

\begin{table}
\caption{Effect of candidate position on selection accuracy across two datasets.}
\centering
\small
\begin{subtable}{\textwidth}
\centering
\caption{Micro-averaged normalized accuracy on ZESHEL for different reranker output list sorting approaches.}
\begin{tabular}{lccccc}
\toprule
\textbf{Model} & \textbf{Reranker} & \textbf{BM25} & \textbf{First} & \textbf{Last} & \textbf{Random} \\
\midrule
\texttt{Magistral-Small-2509} & \underline{81.88} & 81.26 & \textbf{81.91} & 81.60 & 81.56 \\
\texttt{Qwen3-30B-A3B} & \underline{74.42} & 74.34 & \textbf{75.43} & 73.72 & 72.92 \\
\texttt{Qwen3-4B} & \textbf{65.84} & 63.72 & \underline{65.20} & 61.61 & 61.00 \\
\bottomrule
\end{tabular}
\label{tab:positional_bias_zeshel}
\end{subtable}

\vspace{0.8em}

\begin{subtable}{\textwidth}
\centering
\caption{Accuracy on ESCO for different \texttt{UNK} candidate positions.}
\begin{tabular}{llccc}
\toprule
\textbf{Model} & \textbf{Sorting} & \textbf{Acc.} & \textbf{\texttt{UNK} Acc.} & \textbf{Non-\texttt{UNK} Acc.} \\
\midrule
\multirow{3}{*}{\texttt{Magistral-Small-2509}}
& None first & \textbf{26.75} $\pm$ 2.03 & \textbf{5.41} $\pm$ 1.56 & 43.92 $\pm$ 3.06 \\
& None last  & 26.43 $\pm$ 2.03 & 3.94 $\pm$ 1.34 & \textbf{44.51} $\pm$ 3.07 \\
& Random     & 26.48 $\pm$ 2.03 & 4.18 $\pm$ 1.38 & 44.41 $\pm$ 3.07 \\
\midrule
\multirow{3}{*}{\texttt{Qwen3-30B-A3B}}
& None first & \textbf{36.62} $\pm$ 2.21 & \textbf{28.04} $\pm$ 3.09 & 43.52 $\pm$ 3.06 \\
& None last  & 36.51 $\pm$ 2.21 & 27.55 $\pm$ 3.08 & \textbf{43.72} $\pm$ 3.06 \\
& Random     & 35.20 $\pm$ 2.19 & 25.09 $\pm$ 2.99 & 43.32 $\pm$ 3.06 \\
\midrule
\multirow{3}{*}{\texttt{Qwen3-4B}}
& None first & 26.59 $\pm$ 2.03 & 8.36 $\pm$ 1.91 & \textbf{41.25} $\pm$ 3.04 \\
& None last  & \textbf{26.70} $\pm$ 2.03 & \textbf{9.84} $\pm$ 2.05 & 40.26 $\pm$ 3.03 \\
& Random     & 26.32 $\pm$ 2.02 & 8.12 $\pm$ 1.88 & 41.00 $\pm$ 3.04 \\
\midrule
$\emptyset$ (Reranker-only) & - & - & - & \textbf{35.71} $\pm$ 2.96 \\
\bottomrule
\end{tabular}
\label{tab:positional_bias_mhercl}
\end{subtable}

\label{tab:positional_bias_grouped}
\vspace{-\baselineskip}
\end{table}

\section{Runtime and Scalability Results}
\label{sec:runtime}

Figure~\ref{fig:acc_context} shows the selection accuracy as a function of the number of input tokens on ZESHEL.
It shows how the selection accuracy decreases for all three models as the input context length increases. The drop in performance is the least pronounced for the dense \texttt{Magistral-Small-2509} and \texttt{Qwen3-4B} models. The MoE \texttt{Qwen3-30B-A3B} model experiences a significantly steeper drop in performance as the context length increases, eventually achieving worse performance than \texttt{Qwen3-4B} on the longer inputs, despite both models having the same native context window.
Altogether, while LLMs can process significantly longer inputs than smaller language models, these results suggest that LLM-based selection could also benefit from chunking for particularly long documents.
Figure~\ref{fig:acc_thinking_zeshel} reports the selection accuracy as a function of the number of output tokens on ZESHEL. The three curves show a range, around 500 tokens, where the accuracy is the highest. For \texttt{Qwen3-4B} and \texttt{Qwen3-30B-A3B} models, shorter outputs are less accurate, which is not the case for \texttt{Magistral-Small-2509} which does not generate these short outputs. For all three models, the accuracy drops past this range. This is an instance of the typical reasoning LLMs \textit{underthinking} and \textit{overthinking} issue~\cite{suUnderthinkingOverthinkingEmpirical2025a}. Figure~\ref{fig:acc_thinking_unseen} shows a similar ideal range below 500 tokens on another dataset, where all three models achieve the highest, and highly similar, performance, but no \textit{underthinking} phenomena.

Table~\ref{tab:runtime_main} in Section~\ref{sec:analysis} reports the per-mention processing times together with the corresponding accuracies. Table~\ref{tab:tokens} complements them with the thought token usage per self-consistency sample, which directly impacts the runtime of the approach. \texttt{Magistral-Small-2509} generates the longest thoughts on average, with similar values for the other models, and these numbers also vary based on the setting and domain. As discussed in Section~\ref{sec:analysis}, our distillation experiments further show that much of LELA's reasoning capability can be transferred to substantially smaller models, providing an additional path toward reducing inference cost while retaining most of the performance gains.

\begin{figure}
    \centering
    \begin{subfigure}{0.49\linewidth}
        \centering
        \includegraphics[width=\linewidth]{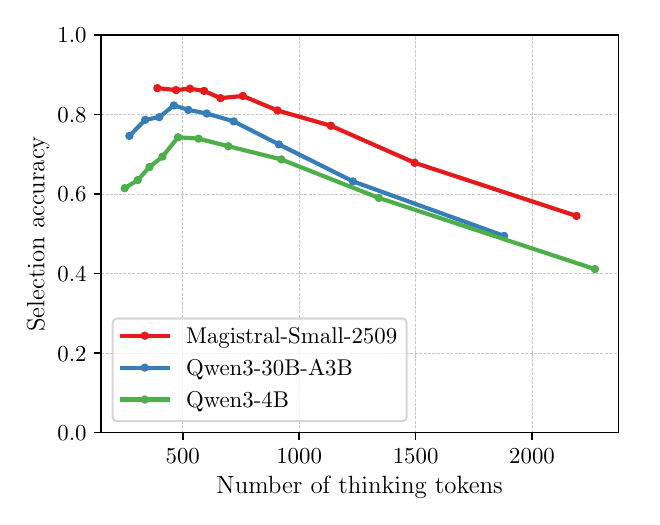}
        \caption{ZESHEL.}
        \label{fig:acc_thinking_zeshel}
    \end{subfigure}
    \hfill
    \begin{subfigure}{0.49\linewidth}
        \centering
        \includegraphics[width=\linewidth]{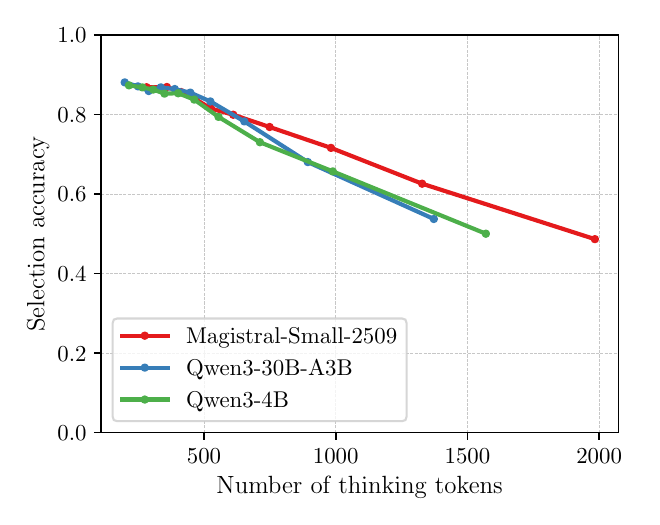}
        \caption{Unseen Mentions.}
        \label{fig:acc_thinking_unseen}
    \end{subfigure}
    \caption{Selection accuracy as a function of the number of generated thinking tokens on two datasets.}
    \label{fig:acc_thinking}
    \vspace{-\baselineskip}
\end{figure}

\begin{figure}
    \centering
    \includegraphics[width=.6\textwidth]{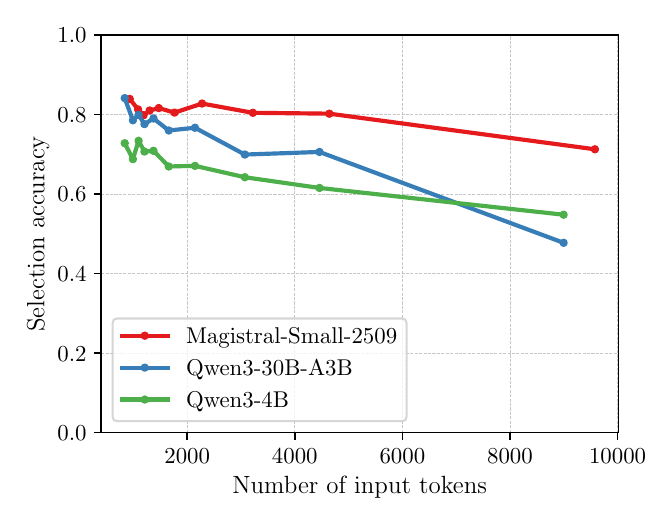}
    \caption{Selection accuracy based on the number of input tokens, on ZESHEL.}
    \label{fig:acc_context}
    \vspace{-\baselineskip}
\end{figure}

\begin{table}
  \centering
  \small
  \caption{Average thought token usage per self-consistency sample, across benchmarks. The ESCO and Unseen Mentions results correspond to the BM25 and prior-based candidate configurations, respectively.}
  \setlength{\tabcolsep}{5pt}
  \begin{tabular}{lcccc}
    \toprule
    \textbf{LELA config} & \textbf{ZESHEL} & \textbf{GLADIS} & \textbf{Unseen M.} & \textbf{ESCO} \\
    \midrule
    w/ Magistral-Small-2509 & 1042 & 1408 & 826 & 1368 \\
    w/ Qwen3-30B-A3B & \phantom{0}947 & \phantom{0}847 & 581 & \phantom{0}882 \\
    w/ Qwen3-4B & 1059 & \phantom{0}847 & 652 & \phantom{0}854 \\
    w/ gpt-oss-20b & \phantom{0}559 & 1172 & 747 & 1491 \\
    \bottomrule
  \end{tabular}
  \label{tab:tokens}
  \vspace{-\baselineskip}
\end{table}

\section{Comparison of Different Pointwise Rerankers}
\label{sec:reranker}
We compared the following rerankers:
\begin{itemize}
    \item BM25, using the outputs from~\cite{logeswaranZeroShotEntityLinking2019}.
    \item The \texttt{Qwen3-Embedding-4B} embedder, computing the similarity between mentions and our entity verbalizations from the candidate list.
    \item The \texttt{Qwen3-4B} LLM with the same prompt and score computation as in~\cite{zhangQwen3EmbeddingAdvancing2025} but without any additional training.
\end{itemize}

\begin{table}
    \centering
    \small
    \caption{Macro-averaged normalized accuracy@10 for different candidate reranking models on ZESHEL.}
    \begin{tabular}{lc}
    \toprule
    Reranker & Acc@10 \\
    \midrule
    \texttt{BM25} & 64.99 \\
    \texttt{Qwen3-Embedding-4B} & 89.94 \\
    \texttt{Qwen3-4B} & 89.12 \\
    \texttt{Qwen3-Reranker-4B} & \textbf{93.57} \\
    \bottomrule
    \end{tabular}
    \vspace{-\baselineskip}
    \label{tab:reranker_metrics}
\end{table}

\noindent The results show that BM25 underperforms compared to the three other approaches, while \texttt{Qwen3-Reranker-4B} achieves the highest recall@10 and \texttt{Qwen3-Embedding-4B} yield competitive results, despite the latter not being provided with the context. Furthermore, Figure~\ref{fig:reranker_grouped} shows the ground truth rank distribution across the four approaches. Interestingly, \texttt{Qwen3-Embedding-4B} assigns the higher score to the ground truth in more than 60\% of the cases, while \texttt{Qwen3-Reranker-4B} does so in less than 30\% of cases, it catches up with the embedder's recall at $k=6$. 
It should be noted that when using models which are highly susceptible to positional bias, such as \texttt{Qwen3-4B} (see Appendix \ref{sec:positional_bias}), the accuracy@k metric is not sufficient to compare different rerankers, as it is invariant to the output candidate position.

\begin{figure}
    \centering
    \begin{subfigure}[b]{0.49\linewidth}
        \centering
        \includegraphics[width=\linewidth]{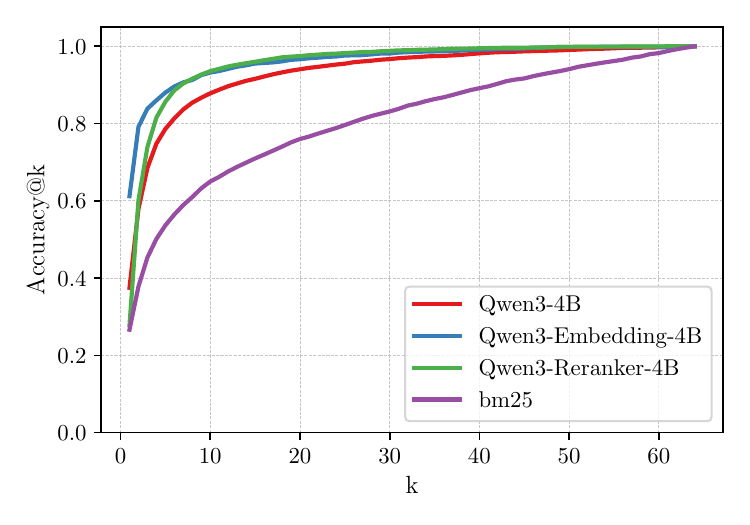}
        \caption{Micro-averaged normalized accuracy@k.}
        \label{fig:reranker_acc_k}
    \end{subfigure}
    \hfill
    \begin{subfigure}[b]{0.49\linewidth}
        \centering
        \includegraphics[width=\linewidth]{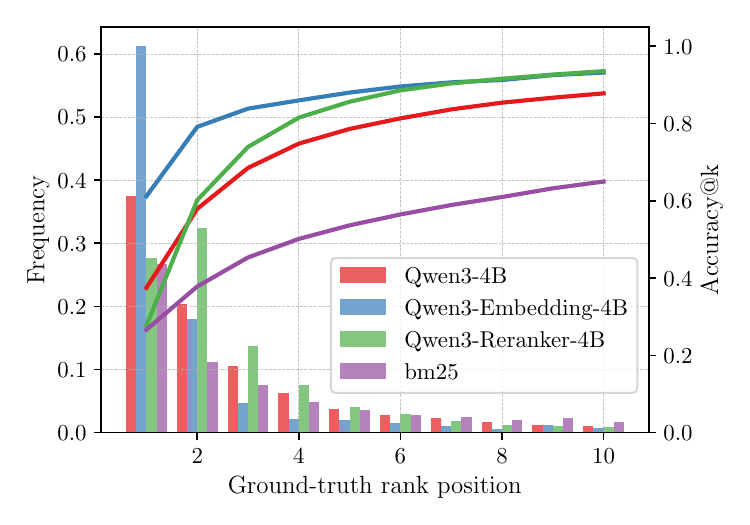}
        \caption{Ground truth position in reranker output histogram.}
        \label{fig:reranker_hist}
    \end{subfigure}
    \caption{Comparison of rerankers on ZESHEL using (a) accuracy@k and (b) ground-truth position histogram.}
    \label{fig:reranker_grouped}
    \vspace{-\baselineskip}
\end{figure}

\section{In-domain Fine-tuning Experiment}
\label{sec:sft_experiment}

This appendix provides full experimental details for the in-domain fine-tuning experiment discussed in Section~\ref{sec:analysis}.

\subsubsection{Setup.} We use the LLM-SFT baseline from Appendix~\ref{sec:baseline} (trained on Wikipedia mentions from ZELDA) as a pre-trained EL system. Our target domain is the Star Trek subdomain of ZESHEL, the largest of the four ZESHEL test domains. We sample 500 mentions where the correct entity is among the candidate set for in-domain fine-tuning, and use all remaining Star Trek mentions for testing.

\subsubsection{Training.} We train for 32 steps with an effective batch size of 16, exposing the model to 16 new in-domain mentions per step. All other hyperparameters match the LLM-SFT baseline (Appendix~\ref{sec:baseline}). We evaluate the model after each training step.

\subsubsection{Comparison.} For each evaluation point, we compare against four LELA variants (\texttt{Qwen3-4B}, \texttt{Qwen3-30B-A3B}, \texttt{gpt-oss-20b}, \texttt{Magistral-Small-2509}) on the same test split. LELA results are computed on the held-out test split (the 500 fine-tuning mentions are excluded from LELA evaluation as well, for fair comparison).

\section{Distillation Experiment}
\label{sec:distillation}

This appendix provides full experimental details for the distillation experiment discussed in Section~\ref{sec:analysis}.

\subsubsection{Data Generation.} We sample 10{,}000 Wikipedia mentions from the ZELDA training split. For each mention, we run \texttt{gpt-oss-20b} (the teacher) with the top-10 candidates returned by \texttt{Qwen3-Reranker-4B}, generating 10 completions per mention. We filter for correctness, keeping only mentions where the teacher's prediction matches the gold entity, and retain a single completion per mention. We parse the reasoning traces from the teacher outputs and convert them into the chat format expected by Qwen3 models, with explicit \texttt{<think>...</think>} tokens followed by an \texttt{"answer": <index>} field.

\subsubsection{Training.} We fine-tune \texttt{Qwen3-0.6B}, \texttt{Qwen3-1.7B}, and \texttt{Qwen3-4B} as students, using LoRA with $r=128$, $\alpha=256$ for one epoch, with an effective batch size of 16 and a learning rate of $1 \times 10^{-5}$. Checkpoint selection is based on loss convergence on a held-out subset of the ZELDA training split.

\subsubsection{Evaluation.} All students are evaluated on ZESHEL using the standard LELA pipeline (top-10 candidates from \texttt{Qwen3-Reranker-4B}, self-consistency over 10 samples). The teacher is evaluated under the same configuration for comparison.

\subsubsection{Note on the True Zero-shot Constraint.} The student models are fine-tuned only on Wikipedia mentions sampled from the ZELDA training split, which contains no overlap with ZESHEL test domains. The distillation procedure therefore preserves LELA's true zero-shot setting on the test benchmarks.

\section{Candidate Retention Funnel}
\label{sec:retention_funnel}

To better understand the contribution of each stage (Retrieval, Reranking, and Selection) to the final performance, we visualize the flow of the ground-truth entity through the LELA pipeline using Sankey diagrams (Figure~\ref{fig:sankey_funnel}).
The diagrams clearly distinguish between retrieval-bound, reranking-bound, and reasoning-bound challenges across different benchmarks:

\begin{itemize}
\item On ZESHEL, the dominant error source is consistently the initial BM25 retrieval. Once the correct entity is retrieved, LELA is highly effective at retaining and selecting it.
\item On GLADIS, the bottleneck shifts to the reranking stage, particularly in specialized domains where the density of similar long forms is high.
\item On WikilinksNED Unseen Mentions (specifically with prior-based candidates), the challenge lies primarily in the final selection, suggesting that the reasoning model struggles to disambiguate between the highly similar entities typical of Wikipedia.
\end{itemize}

\noindent A key finding is the high retention rate of the reranking stage across most datasets. For example, on the ZESHEL/Forgotten Realms domain, the reranker preserves 98\% of the recoverable recall (977 retained vs. 22 lost), while compressing the candidate list by a factor of 6.4 (from 64 to 10). These results demonstrate the robustness of our choice of $k=10$, validating that aggressive filtering does not compromise the upper bound of the system.
However, the results also highlight domain-dependent variations. While the reranker achieves a 90.5\% retention rate on the GLADIS/General domain, this drops to 63.2\% on the GLADIS/Scientific domain. This significant drop indicates that scientific acronyms pose a harder challenge for the reranker, likely due to the fine-grained technical distinctions required to separate valid candidates. Crucially, the Selection LLM proves significantly more robust in this technical setting, successfully identifying the correct entity among the top-10 candidates in 74.1\% of the cases passed to it (6593 out of 8894).

\begin{figure}
\centering

\begin{subfigure}{0.45\textwidth}
  \centering
  \includegraphics[width=\linewidth]{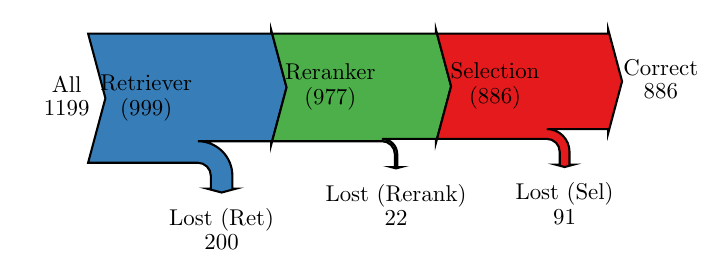}
  \caption{ZESHEL/Forgotten Realms}
\end{subfigure}\hfill
\begin{subfigure}{0.45\textwidth}
  \centering
  \includegraphics[width=\linewidth]{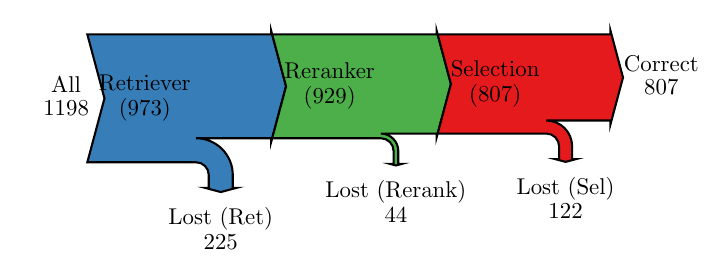}
  \caption{ZESHEL/Lego}
\end{subfigure}\hfill

\vspace{1em}

\begin{subfigure}{0.45\textwidth}
  \centering
  \includegraphics[width=\linewidth]{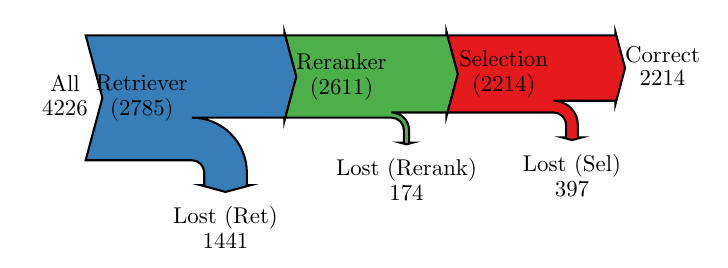}
  \caption{ZESHEL/Star Trek}
\end{subfigure}\hfill
\begin{subfigure}{0.45\textwidth}
  \centering
  \includegraphics[width=\linewidth]{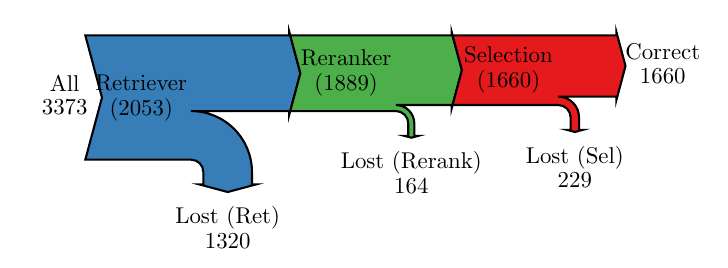}
  \caption{ZESHEL/YuGiOh}
\end{subfigure}

\vspace{1em}

\begin{subfigure}{0.45\textwidth}
  \centering
  \includegraphics[width=\linewidth]{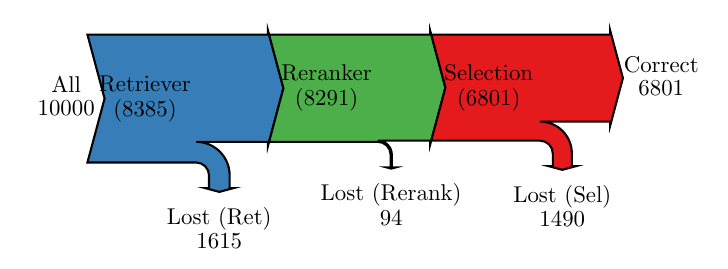}
  \caption{WikilinksNED Unseen Mentions}
\end{subfigure}\hfill
\begin{subfigure}{0.45\textwidth}
  \centering
  \includegraphics[width=\linewidth]{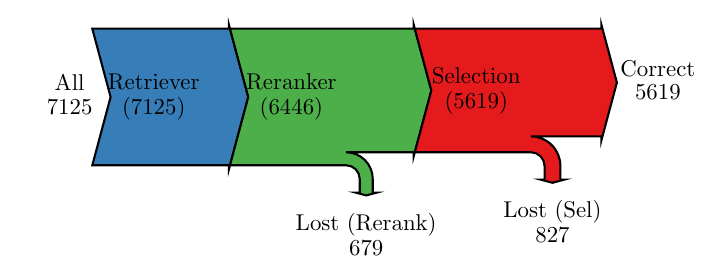}
  \caption{GLADIS/General}
\end{subfigure}\hfill

\vspace{1em}

\begin{subfigure}{0.45\textwidth}
  \centering
  \includegraphics[width=\linewidth]{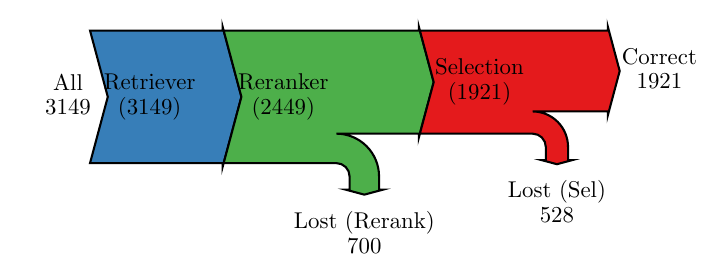}
  \caption{GLADIS/Biomedical}
\end{subfigure}\hfill
\begin{subfigure}{0.45\textwidth}
  \includegraphics[width=\linewidth]{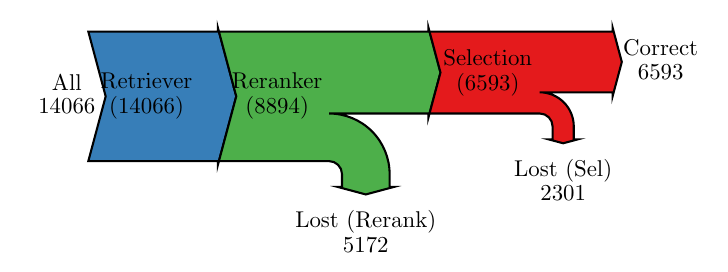}
  \caption{GLADIS/Scientific}
\end{subfigure}

\caption{Candidate retention funnel across datasets, using \texttt{Magistral-Small-2509}. The diagrams visualize the attrition of the ground-truth entity at each stage of the LELA pipeline: (1) from the total dataset to the Retriever, (2) to the Reranker, and (3) to the final Selection.}
\label{fig:sankey_funnel}
\vspace{-\baselineskip}
\end{figure}

\FloatBarrier

\section{Examples}
\label{sec:examples}
Figures~\ref{fig:none_example}, \ref{fig:reranker_example_success}, \ref{fig:reranker_example_fail}, \ref{fig:description_example}, \ref{fig:selection_example}, and \ref{fig:model_example} show examples of reasoning traces in the selection phase of LELA.

\begin{figure}
    \small
    \raggedright
    \begin{framed}
    \textbf{Input sentence and mention:}
        \textbf{[Develop tools]} for our internal teams ( sales support operations finance ) to help run the business more efficiently.
        
    \textbf{LELA:} <think> The task is to determine which of the candidates best fits the mention "[Develop tools]" in the context of developing tools for internal teams to improve business efficiency [\ldots] Option 5 and 6 are about business process modelling [\ldots] Option 8 is software frameworks [\ldots] Option 10 is about content development tools, which doesn't fit [\ldots] \textcolor{blue}{option 5 and 6 are about business process modelling, which might be considered tools. But the original mention is more general. Since the options don't have a general "develop tools for internal efficiency" category, perhaps 0 is the answer [\ldots] But the original text says "tools" not "models". So perhaps 0 is the correct choice here.}</think>
    \textbf{"answer": 0}. \textcolor{green!60!black}{(\cmark)}
    \end{framed}
    \caption{LLM selection example from ESCO for the \texttt{UNK} entity. The model correctly chooses the \texttt{None} candidate as shown in \textcolor{blue}{blue} instead of the other strong candidates.}
    \label{fig:none_example}
    \vspace{-\baselineskip}
\end{figure}

\begin{figure}
    \small
    \raggedright
    \begin{framed}
    \textbf{Input sentence and mention:}
        The Golden Rule is arguably the most essential basis for the modern concept of human rights , in which each individual has a \textbf{[right]} to just treatment, and a reciprocal responsibility to ensure justice for others. [ 4 ] A key element of the Golden Rule is
        
    \textbf{LELA w/o LLM selection:} \textcolor{red}{\textbf{Right to keep and bear arms}: The right to keep and bear arms (often referred to as the right to bear arms) is a legal right for people to possess weapons (arms) for the preservation of life, liberty, and property} \textcolor{red}{(\xmark)} \\    
        
    \textbf{LELA:} <think> The marked mention is [right], in the context of human rights and the Golden Rule [\ldots] Option 4 is Human rights, but the question is about the specific entity that fits [right] in the sentence [\ldots] Option 6 is "Rights" in general, described as principles of freedom or entitlement. \textcolor{blue}{The sentence uses "right" in a general sense, not a specific type like the right to bear arms or civil rights.} [\ldots] The Golden Rule is linked to human rights, but the specific mention here is "right" as a general concept. So I think the answer is 6.</think>
    \textbf{"answer": 6}. \textcolor{green!60!black}{(\cmark)}
    \end{framed}
    \caption{Reranker and LLM selection example from WikilinksNED Unseen Mentions. When using the selection LLM, as shown in \textcolor{blue}{blue}, the model is able to select the correct candidate, unlike the reranker which ranks a wrong candidate (shown in \textcolor{red}{red}) first.}
    \label{fig:reranker_example_success}
    \vspace{-\baselineskip}
\end{figure}

\begin{figure}
    \small
    \raggedright
    \begin{framed}
    \textbf{Input sentence and mention:}
Fourvière , annual festival in June and August. The Biennale d'art contemporain . The Nuits Sonores . Lille ( French pronunciation: [lil] ; \textbf{[Dutch]} : Rijsel) is a city in northern France . It is the principal city of the Lille Métropole , the fourth-largest metropolitan area in the
        
    \textbf{LELA w/o LLM selection:} \textcolor{green!60!black}{\textbf{Dutch Language}: Dutch ( ) is a West Germanic language spoken by about 25 million people as a first language and 5 million as a second language. (\cmark)} \\    
        
    \textbf{LELA:} <think> Okay, let's tackle this problem. \textcolor{red}{The mention in the text is "Lille,"} but the candidates don't include Lille directly [\ldots] So perhaps the correct answer is related to the Low Countries (option 2), since Lille is in northern France, near the border with the Low Countries [\ldots] But I'm not entirely sure [\ldots] So the answer is 2. </think> \textbf{"answer": 2} \textcolor{red}{(\xmark)}

    \end{framed}
    \caption{LLM selection failure example from WikilinksNED Unseen Mentions. The reranker ranks the correct candidate first. However, the selection LLM considers the wrong mention, as shown in \textcolor{red}{red}, overthinks, and selects a wrong candidate.}
    \label{fig:reranker_example_fail}
\end{figure}

\begin{figure}
    \small
    \raggedright
    \begin{framed}
    \textbf{Input sentence and mention:}
        Ann Lee is a minifigure introduced in the Monster Fighters theme [\ldots] Ann Lee is a Monster Fighter, proficient in the ancient art of broomstickajitzu. Her monster arch-enemy are the Ghosts, even though she is shown in videos with the \textbf{[Mummy]}.

    \textbf{LELA w/o entity descriptions:} <think> The mention is [Mummy] in the context of Ann Lee, a Monster Fighters minifigure [\ldots] The list includes various Mummy-related entries: Mummy (Adventurers), The Mummy, Mummy (Scooby-Doo), Part numbers, and others [\ldots] Since the original text is about a minifigure, the most likely answer would be "Mummy (Minifigures)" which is option 6 [\ldots] \textcolor{red}{Alternatively, maybe "The Mummy" (option 2 or 8). But considering the list, option 6 is more specific to minifigures.} So I think the answer is 6. </think> \textbf{"answer": 6} \textcolor{red}{(\xmark)}\\    
        
    \textbf{LELA:} <think> The mention in question is [Mummy], in the context of Ann Lee, a Monster Fighter minifigure [\ldots] Option 2 is "The Mummy King is a Monster Fighters minifigure..." and option 9 is "9462 The Mummy" which is a set [\ldots] Since the text refers to the Mummy as a character, it's more likely referring to the minifigure. So option 2 [\ldots] Option 1 is Mummy (Adventurers), from a different game. Option 3 is Scooby-Doo, a different theme. \textcolor{blue}{Option 5 and 6 are other minifigures but not from Monster Fighters.} Option 8 is a Studios minifigure, but the text mentions Monster Fighters. So yes, option 2 is the correct one. </think> \textbf{"answer": 2} \textcolor{green!60!black}{(\cmark)}

    \end{framed}
    \caption{LLM selection example from ZESHEL/Lego when using or omitting entity descriptions. Without descriptions, the LLM selects the entity with the most relevant title given the context, as shown in \textcolor{red}{red}. Using entity descriptions, it excludes this wrong candidate, as shown in \textcolor{blue}{blue}.}
    \label{fig:description_example}
\end{figure}

\begin{figure}
    \small
    \raggedright
    \begin{framed}
    \textbf{Input sentence and mention:}
        The Dog People [\ldots] were primitive, savage, nomadic tribesmen who lived in the deserts of northern Maztica [\ldots] Religion. The Dog People practiced a shamanistic religion, and revered all of the gods of the , especially \textbf{[Maztica]} and Tezca. [\ldots]
        
    \textbf{LELA ($k=64$):} <think> The mention is "[Maztica]". Candidate 1 is "Maztica" itself, described as a continent west of Faerûn. The text mentions that the Dog People lived in the deserts of northern Maztica, so this seems like a strong candidate. \textcolor{red}{Other candidates like "Maztica trilogy" (2), "Maztican" (3), "Kukul" (4), etc., don't fit because they refer to different things---like a book series, a term for inhabitants, a god, etc.} [\ldots] So the correct choice is candidate 1, "Maztica". </think> \textbf{"answer": 1} \textcolor{red}{(\xmark)} \\    
        
    \textbf{LELA ($k=10$):} <think> The actual sentence is: "The Dog People practiced a shamanistic religion, and revered all of the gods of the, especially [Maztica] and Tezca." So the mention is in the context of gods. \textcolor{blue}{Candidate 1 is Maztica (the continent). Candidate 3 is Maztican (inhabitants). Candidate 7 is Maztica (goddess) [\ldots] But the mention here is [Maztica], and the text says they revered the gods of the, especially Maztica and Tezca. So Maztica here is likely a deity, not the continent or the people [\ldots] Therefore, the correct candidate should be 7.} </think> \textbf{"answer": 7}. \textcolor{green!60!black}{(\cmark)}
    \end{framed}
    \caption{LLM selection example from ZESHEL/Forgotten Realms for $k=10$ and $k=64$ (all retrieved candidates). When considering all retrieved candidates, the model overlooks the seventh candidate, which is the correct one, and takes shortcuts, as highlighted in \textcolor{red}{red}. When only considering ten candidates, it successfully considers the seventh candidate, among other options, as highlighted in \textcolor{blue}{blue}, and selects it.}
    \label{fig:selection_example}
\end{figure}

\begin{figure}
    \small
    \raggedright
    \begin{framed}
    \textbf{Input sentence and mention:}
        hydrogen is less than 25\%. [ 47 ] [ 48 ] [ 49 ] [ edit ] Production For more details on this topic, see \textbf{[Hydrogen production]}. The molecular hydrogen needed as an on-board fuel for hydrogen vehicles can be obtained through many thermochemical methods utilizing natural gas, coal (by
        
    \textbf{LELA (\texttt{Qwen3-4B}):} <think> The text mentions producing molecular hydrogen for hydrogen vehicles through thermochemical methods using natural gas and coal. The mention of "natural gas" and "coal" suggests methods of production [\ldots] "Steam reforming" (option 2) is a method that uses natural gas and water to produce hydrogen [\ldots] \textcolor{red}{The other options like "Hydrogen production" (1) is a general term, but the text is discussing specific methods.} So the answer should be option 2. </think> \textbf{"answer": 2} \textcolor{red}{(\xmark)} \\    
        
    \textbf{LELA (\texttt{Qwen3-30B-A3B}):} <think> The marked mention is [Hydrogen production] [\ldots] \textcolor{blue}{The candidate list includes "Hydrogen production" as entry 1. The text explicitly says "see [Hydrogen production]" which is a direct reference. So even though there are other entries like steam reforming (6) or copper-chlorine cycle (1), the context here is pointing towards the general topic of hydrogen production.} So the answer should be 1. </think> \textbf{"answer": 1}. \textcolor{green!60!black}{(\cmark)}
    \end{framed}
    \caption{LLM selection example from WikilinksNED Unseen Mentions for two selection LLMs. \texttt{Qwen3-4B} chooses the wrong candidate, following the thoughts in \textcolor{red}{red}, while \texttt{Qwen3-30B-A3B} chooses the right answer, following the \textcolor{blue}{blue} thoughts.}
    \label{fig:model_example}
\end{figure}

\end{document}